\documentclass[preprint,journal]{vgtc}       

\usepackage{mathptmx}
\usepackage{graphicx}
\usepackage{times}
{}
{}
{}
\usepackage{lscape}
\usepackage{tabularx}
\usepackage{pbox}
\usepackage{multirow}
\usepackage{array, makecell} %
\usepackage[T1]{fontenc}
\usepackage[utf8]{inputenc}
\usepackage[bookmarks,backref=true,linkcolor=black]{hyperref} 
\hypersetup{
  pdfauthor = {},
  pdftitle = {},
  pdfsubject = {},
  pdfkeywords = {},
  colorlinks=true,
  linkcolor= black,
  citecolor= black,
  pageanchor=true,
  urlcolor = black,
  plainpages = false,
  linktocpage
}

\onlineid{0}

\vgtccategory{Research}

\vgtcinsertpkg

\preprinttext{This manuscript has been accepted for publication in IET Networks.}

\title{Artificial Intelligence Enabled Software Defined Networking: A Comprehensive Overview}

\author{Majd Latah and Levent Toker}

\authorfooter{
\item
 Majd Latah is with Ozyegin University. E-mail: majd.latah@ozu.edu.tr.
\item
 Levent Toker is with Ege University. E-mail: levent.toker@ege.edu.tr.
}

\abstract{Software defined networking (SDN) represents a promising networking architecture that combines central management and network programmability. SDN separates the control plane from the data plane and moves the network management to a central point, called the controller, that can be programmed and used as the brain of the network. Recently, the research community has showed an increased tendency to benefit from the recent advancements in the artificial intelligence (AI) field to provide learning abilities and better decision making in SDN. In this study, we provide a detailed overview of the recent efforts to include AI in SDN. Our study showed that the research efforts focused on three main sub-fields of AI namely: machine learning, meta-heuristics and fuzzy inference systems. Accordingly, in this work we investigate their different application areas and potential use, as well as the improvements achieved by including AI-based techniques in the SDN paradigm.
} 

\CCScatlist{ 
 \CCScat{K.6.1}{Management of Computing and Information Systems}%
{Project and People Management}{Life Cycle};
 \CCScat{K.7.m}{The Computing Profession}{Miscellaneous}{Ethics}
}

\begin{document}



\maketitle

\section{Introduction}
Software-defined networking (SDN) adopts the concept of programmable networks by using a logically centralized management, which represents a simplified solution for complex tasks such as traffic engineering [1], network optimization [2] and orchestration [3]. Furthermore, dealing with modern network applications requires more scalable architecture which should be able to provide reliable and sufficient services based on a specific traffic type [4]. This can be achieved with the SDN architecture, which maintains a global view of network states and provides a flow-level control of the underlying layers [4]. This idea caused a dramatical change in the way how networks are designed and managed [5, 6]. In addition, the SDN architecture allows the involvement of third parties in the design and deployment of modern network applications [6]. Ethane project [7] formed the foundation of today's SDN by presenting a new networking paradigm, in which a centralized controller is used for flow-level policy management and security purposes in enterprise networks. \\

The SDN paradigm separates the control plane from the data plane, which results in achieving much faster and dynamic approach in compared with a conventional network architecture [8]. The control plane can be split into several virtual networks where each one implements a different policy [8]. As a result, this paradigm can be viewed as a tool allows addressing various issues in networking from another perspective [5] and can be used also to fulfil the requirements of new technologies such as internet of things (IoT) and 5G [9]. The adoption of SDN paradigm, however, strongly depends on its success in reaching an appropriate solution for the problems, which cannot be solved by the traditional networking protocols and architectures [10]. Some large companies such as Microsoft and Google have already started using the SDN paradigm for their own data centres [11, 12]. Artificial Intelligence (AI), on the other hand, reveals a huge potential in SDN innovation. Our previous study [13] focused on highlighting the first efforts to integrate AI in SDN. In this work, however, we provide a thorough overview of the research efforts in this area to a gain deeper insight into the significant role of AI in SDN paradigm.

\section{SDN Architecture}
As mentioned previously, SDN promotes innovation by introducing the concept of centralized programmable control of the data plane, which facilitates the development of new network services and protocols [4]. The SDN architecture is designed based on the idea of the separation between control and data planes (see Fig. 1).

\begin{figure*}[ht!]
\centering
\includegraphics[width=140mm]{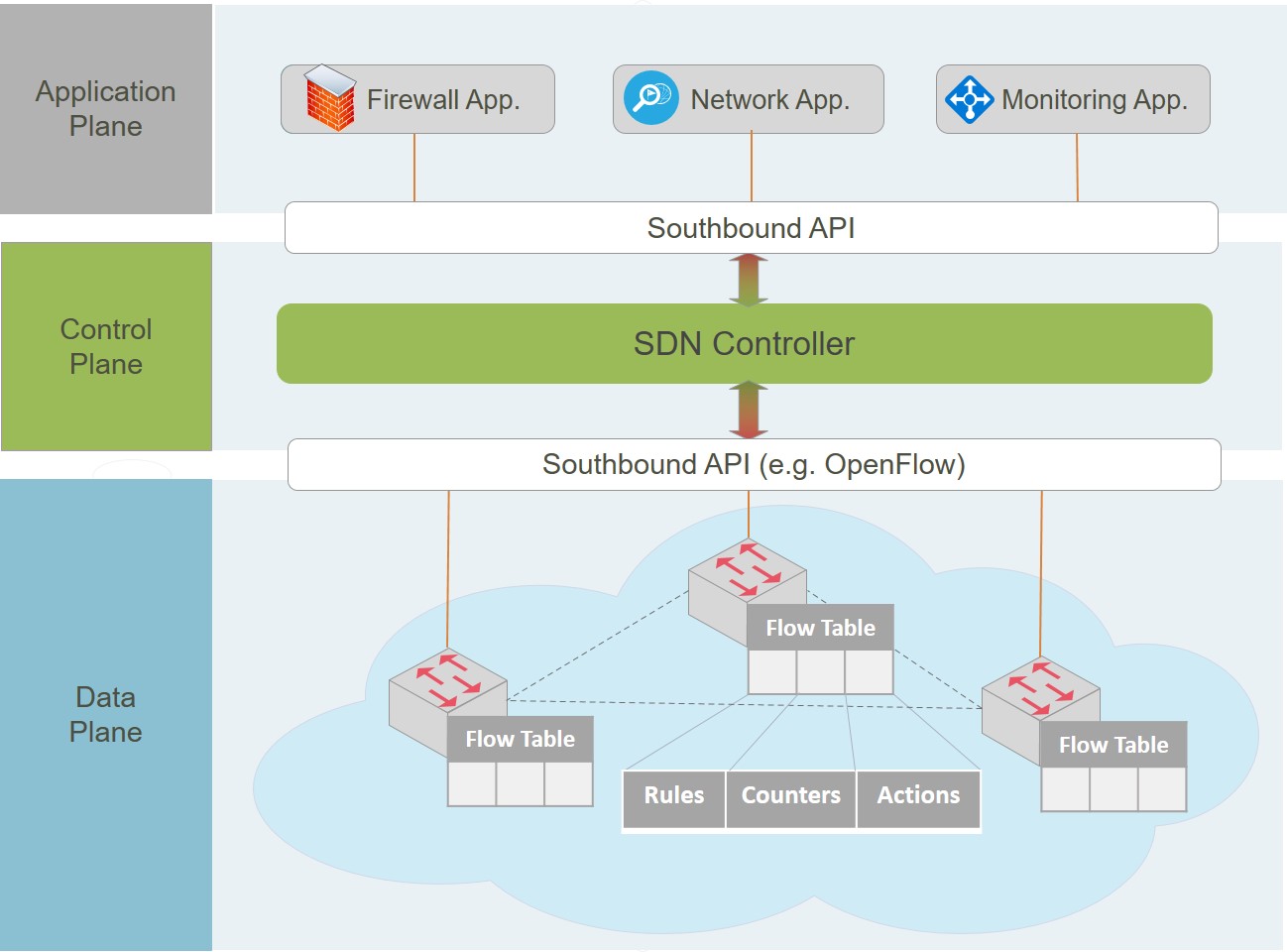}
\caption{A basic SDN architecture \label{overflow}}
\end{figure*}

The first attempt was network control point (NCP) [14], which employed the separation concept to enhance the control of AT\&T's telephone network, whereas recent contributions such as Ethane [7] and SANE [15] applied the same concept for Ethernet networks [16]. The applications of SDN reside in the application plane of SDN architecture where the northbound application programming interface (API) provides the commutation between the application and control planes [8], which allows implementing a set of network services such as traffic engineering, intrusion detection, quality of service (QoS), firewall and monitoring applications [4]. Northbound API allows developers to write their own applications without the need for a detailed knowledge of the controller functions or understanding how the data plane works. It is worth mentioning that several SDN controllers provide their own northbound APIs [16]. 

The communication between control and data planes is provided using a southbound API such as forwarding and control element separation (ForCES) [17], open vSwitch database (OVSDB) [18], protocol oblivious forwarding (POF) [19], OpenState [20], OpenFlow (OF) [21] and OpFlex [22], which enables exchanging control messages with forwarding elements (e.g., OF-enabled switches). 
As shown in Figure 1, each OF-enabled switch adopts a flow-based decision making logic determined by the so-called SDN controller, which is responsible for preparing the forwarding tables of each switch [8]. A typical OF-enabled switch has a pipeline of flow tables, which consist of flow entries, each of which has three parts: (1) matching rules, which are used for matching incoming packets (2) counters that maintain statistics of matched flows and (3) actions or instructions, which can be configured proactively or reactively to be executed upon a match [6, 14]. The forwarding elements (i.e. OF-enabled switches) can be implemented in either software or hardware. Some software switches such as Open vSwitch have a great potential for providing a solution for data centres and virtual networks [16]. On the other hand, other APIs [10, 23] are proposed for a specific purpose (e.g, VOIP applications and inter-domain routing, not to mention various SDN programming languages such as Procera [24], NetCore [25] and Frenetic [26], which provide high-level APIs that can be used to develop different SDN applications in more flexible and functional manner.

\section{OpenFlow Protocol}
OpenFlow (OF) is considered the most commonly used southbound API in SDN, which is being continuously developed and standardized by open networking foundation (ONF) [6]. OF provides an abstraction layer that enables the SDN controller to securely communicate with OF-enabled forwarding elements [6]. OpenFlow has become the de-facto standard for southbound APIs used in SDNs [6] and therefore, in this study, we mainly focus on OF-based SDNs. OF-based forwarding devices have been developed to coexist together with conventional Ethernet devices [16]. Hybrid switches, on the other hand, reveal new possibilities by including both OF and non-OF ports [6]. As we mentioned previously, a set of control messages can be sent by the controller to prepare and update a particular switch's flow tables. A typical OF-enabled switch handles new coming packets based on its flow table. Figure 2, shows the fields of matching rules part in OF version 1.0.0. A table-miss occurs when a new packet does not match any of the flow table entries. In this case, the switch may either drop the packet or forward it to the corresponding controller using OF protocol [6].

\begin{figure*}
\centering
\includegraphics[width=150mm]{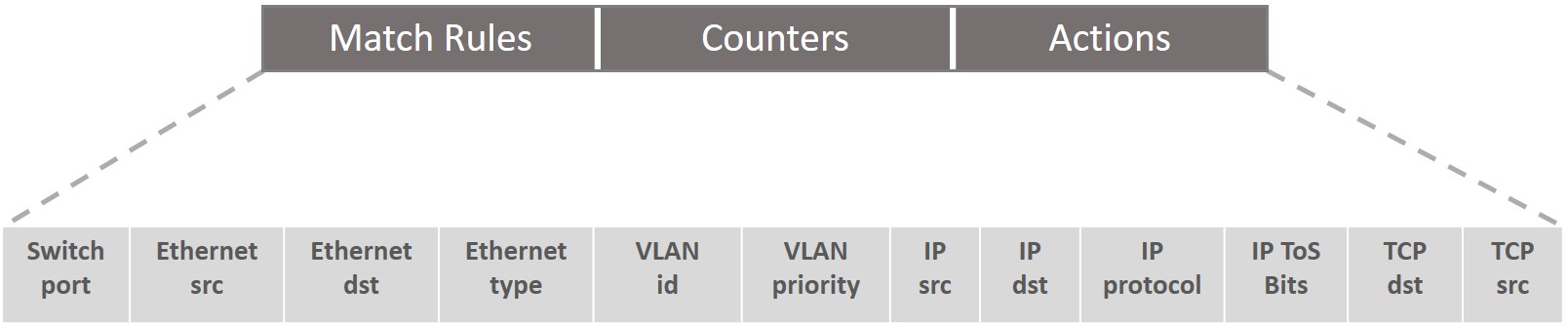}
\caption{Structure of OpenFlow V1.0.0 matching rules\label{overflow}}
\end{figure*}

It is worth mentioning that the identity-based access control of the Ethane project [7] became the first specification of OF switches. After the release of the first OF specification in Spring 2008, many vendors such as HP, NEC, Cisco and Juniper were able to build their first OF-enabled hardware switches, where NOX [27] was the only available controller at that period of time [28]. Recently, different versions of OF protocol have been introduced to add more flexible and reliable capabilities by including multiple flow tables, enhanced matching/action abilities, optical ports, group tables, meter tables and synchronized tables [4, 29]. More details concerning various versions of OF specification can be found in [30]. In addition, there are many available OF controllers, such as POX [31], Beacon [32], OpenDayLight [33], Floodlight [34], and Ryu [35].

\section{Artificial Intelligence}
Artificial Intelligence (AI) is a rapidly growing field that includes a wide range of sub-fields, including knowledge representation, reasoning, planning, decision making, optimization, machine learning (ML) and meta-heuristic algorithms. Turing Test [36] provides a fulfilling definition of intelligence, where a computer has to answer some questions written by a human interrogator. Accordingly, the computer passes the test if the interrogator cannot tell whether the answer was written by a human or a machine. In order to pass Turing Test, the computer need to include advanced capabilities such as natural language processing, knowledge representation, automated reasoning, machine learning and computer vision [36]. AI research started later in the mid-1950s, where a summer workshop organized by Martin Minsky and Claude Shannon at Dartmouth College resulted in the birth of the field of AI [37]. The first contribution, however, was made in 1943 by McCulloch and Pitts, when they proposed the first model for artificial neural networks, in which each neuron has a binary output (-1,+1) with a sign activation function [37]. Adoption of AI approaches increased, thanks to key contributions led to the emergence of new sub-fields such as expert systems, fuzzy logic and evolutionary computation. Further efforts have fuelled the research in AI field by refining the existing methods and proposing brand new hybrid intelligent approaches. Machine learning, meta-heuristic algorithms and fuzzy inference systems are widely used in SDN paradigm therefore we provide an introduction regarding these approaches.

\subsection{Machine Learning}
An intelligent machine learns from experience (i.e., learns from the data available in its environment) and uses it to improve the overall performance [36,37]. In this context, learning methods fall into four groups:

\subsubsection{Supervised learning}\mbox{} \\
Supervised learning methods are provided with a pre-defined knowledge. For instance, a training dataset that consists of input-output pairs, where the system learns a function that maps a given input to an appropriate output [36]. This approach requires having a dataset that represents the system under consideration and can be used to estimate the performance of the selected method [38].

\paragraph{Artificial neural networks}\mbox{} \\
Neural networks are mainly inspired from biological learning systems such as biological neurons in human brain [37]. Artificial neural networks have many advantages. First, they can adjust themselves to the data without explicitly specifying a functional or distribution for representing the underlying model [39]. Second, neural networks form a universal functional approximator, which can approximate any function [39]. Third, neural networks are non-linear models, which gives them the flexibility to represent and model complex relationships [39]. The feed-forward multilayer networks or multilayer perceptrons (MLPs) are the most common used neural network classifiers. MLPs are mainly trained with supervised training algorithms. Neural networks are subject to over-fitting when we use too many parameters in our model [36]. We also need to find the best network structure to achieve a good performance [36].

\paragraph{Support vector machines (SVM)}\mbox{} \\
This algorithm finds linear separators that maximize the margin between two different classes in order to provide a better generalization of the classifier. Kernel methods can be used to transform the input data into a high-dimensional space in order to deal with linearly non-separable cases [36]. SVMs can represent complex functions and show robustness against over-fitting [36].

\paragraph{Decision trees}\mbox{} \\
Decision trees (DT) are very successful in classification problems. DT can describe a data set by a tree-like structure [37]. The input and output data can be discrete or continuous. Decision trees can represent all Boolean functions. A decision tree performs a sequence of tests, where each internal node in the tree corresponds to a test of one of the input attributes [36]. Interpretability (i.e. the ability to understand the reason for the output of the learning algorithm) is one of main advantages of DT, since this approach is very natural for humans [36]. DT, however, sufferers from over-fitting where it may lead to a large tree when there is no pattern to be found in the input data [36].

\paragraph{Ensemble methods}\mbox{} \\
The ensemble methods combine predictions of different approaches (by weighted or unweighted voting) and mainly used for improving the performance of learning algorithms [36]. Bagging, represents the first effective method used for increasing the accuracy by creating an improved composite classifier that combines different outputs of learned classifiers into a single output. Each one of these classifiers is trained by instances generated by random sampling with replacement from the original data set [36,40]. In boosting, unlike bagging, each classifier is influenced by the performance of the previous classifier and tries to pay more attention to the errors made by the previous classifier [40]. For more details, we refer the reader to [40].

\paragraph{Supervised deep learning}\mbox{} \\
Deep learning provides a general-purpose multi-level representation-learning approach [41]. In representation learning a machine can learn to automatically discover the representations required for classification or detection task based merely on raw data, whereas conventional machine-learning techniques cannot deal natural data in their raw form [41]. Multiple levels of representation allows transforming the representation from low level into a higher abstract one. An enough number of these transformations allows learning more complex functions [41]. Deep learning techniques [41-42], have achieved better performance compared to the traditional algorithms used for many machine learning tasks such as speech recognition, intrusion detection, objection detection and natural language understanding. Deep learning models are categorized into three groups, namely: 1) generative, 2) discriminative, and 3) hybrid models. Discriminative models mainly use supervised learning approaches, whereas generative models employ unsupervised learning approaches. Hybrid models, on the other hand, make use of both discriminative and generative models [42,43]. In this paper, the term deep neural networks (DNN) refers to deep feed-forward multilayer networks or multilayer perceptrons (MLPs). Other important two supervised models used in deep learning are recurrent neural networks and convolutional neural networks.

\subparagraph{Recurrent neural networks}\mbox{} \\
Recurrent neural network (RNN) is an extension of feed-forward neural networks that addresses sequential (e.g., speech or text) or time-series problems [43, 44]. Unlike traditional feed forward neural network, the output of an RNN depends on the previous computations [44], this is the basic reason why it is called an recurrent neural network [44]. The Back-propagation Through Time (BPTT) algorithm is mainly employed for the training stage of an RNN. However, the conventional RNN encounters vanishing/exploding gradient problems. Long Short Term Memory (LSTM) networks and Gated Recurrent Units (GRUs) were proposed to cope with this problem [44].

\subparagraph{Convolutional neural networks}\mbox{} \\
Convolutional neural networks (ConvNets) were proposed to process and deal with the data that comes in the form of multiple arrays [41] such as images. ConvNets are used in feature learning for large-scale image classification. A typical CovNet, consists of three layers: 1) convolutional layer, 2) sub-sampling layer (pooling layer) and 3) fully-connected layer [45]. In the convolutional layer, a filtering operation performed by a feature map (i.e. discrete convolution) is used to attain the weight sharing, whereas the sub-sampling is employed in the pooling layer for dimensionality reduction [41,45].

\subsubsection{Unsupervised learning}\mbox{} \\
Unsupervised learning methods are provided without a pre-defined knowledge (i.e., having an unlabelled data) [36]. Therefore, the system mainly focuses on finding specific patterns in the input. An example of the unsupervised learning approach is clustering, which is used for detecting useful clusters in the input data based on similar properties defined by a proper distance metric such as Euclidian, Jaccard, and cosine distance metrics [36,38].

\paragraph{K-means clustering}\mbox{} \\
K-means [46] is one of the well-known clustering approaches. A prior knowledge of the parameter \textit{k}, which indicates 
the number of the resulted clusters, is needed for this algorithm. Each data point will be assigned to the nearest centroid of each cluster. K-means minimizes an objective function that represents the distance between the data points and their corresponding centroids [47]. The process of updating the centroids, based on their assigned data points, will be repeated until the centroids remain the same or no point changes. K-means depends mostly on the initial set of clusters. Therefore an inappropriate choice of k may resulted in poor results [48]. Moreover, fuzzy-C-Means (FCM) [49] clustering, which is also known as soft K-means, allows each data point to belong to more than one clusters. In other words, a data point can belong to all clusters with different degree of membership.

\paragraph{Self-organizing maps (SOM)}\mbox{} \\
Self-organizing maps (SOM) [50-51], is a well-known unsupervised learning approach in artificial neural networks, which
maps high-dimensional distribution to a low-dimensional representation called a SOM map [52]. SOMs have proven to be successful in various pattern recognition tasks including very noisy signals [51]. The training process in SOM, builds and reorganizes the map using input data. Thereafter, it classifies a new input vector based on finding its winning neuron or node in the map [52].

\paragraph{Hidden Markov model (HMM)}\mbox{} \\
Hidden Markov Model(HMM) [53], represents a statistical model, where the system under development is assumed to be a Markov process with unobserved (hidden) states [54]. The Markov process is the random process, which fits for the Markov assumption, where Markov assumption is that the probability of one state depends only on the previous state [54]. The HMM specifies five entities in the model which are: 1) the set of states, 2) the output alphabet, 3) the initial probability state, 4) transition probabilities, and 5) the observation probability. The parameters of HMM can be trained in supervised or unsupervised manner [54]. Baum-Welch algorithm [53] is considered to be the most common used unsupervised algorithm in HMM [54].

\paragraph{Restricted Boltzmann machine (RBM)}\mbox{} \\
An RBM represents a stochastic ANN that consists of two layers: input layer and hidden layer [43]. The restriction of RBMs in compared with basic Boltzmann machine is that the connectivity of the neurons, where each neuron in the input layer is connected to all of the hidden neurons and vice versa, but there is no connection between any two neurons in the same layer [43]. Moreover, the bias unit is connected to all of the visible and hidden neurons. RMBs are an essential component in deep belief networks (DBNs) and can be used for feature extraction [43].

\paragraph{Unsupervised deep learning approaches}\mbox{} \\
As we mentioned previously, deep architectures are classified into: 1) generative, 2) discriminative, and 3) hybrid models.
We mentioned also that the generative models employ unsupervised learning approaches to characterize the high-order correlation properties of the input data [42]. The generative models need an unsupervised pre-training stage to extract the structures in the input data. They also need an additional top layer to perform the discriminative task [42]. In this paper, we discuss two deep generative models namely stacked auto-encoder (SAE) and deep belief network (DBN), due to the fact that they were already employed in SDNs [55,56-58].

\subparagraph{Stacked auto-encoder (SAE)}\mbox{} \\
Auto-encoder (AE) are suitable for feature extraction and dimensionality reduction [43]. A basic auto-encoder has two stages: 1) encoding and 2) decoding stage [45]. The first stage receives the input data and transforms it to a new representation, called a code or latent variable, whereas the second stage receives the generated code at the first stage and reconstruct the original input data [43]. The training procedure aims at minimizing the reconstruction error [43]. An SAE is trained by a two-stage approach. The pre-training stage includes training the initial parameters in a greedy layer-wise unsupervised style, whereas the next fine-tuning stage makes use of a supervised approach to fine-tune the parameters of the model with respect to the labelled instances by adding a softmax layer on the top layer [45].

\subparagraph{Deep belief network (DBN)}\mbox{} \\
Deep belief network (DBN) is a type of generative ANNs that represents the first successful deep learning model in which several restricted Boltzmann machines (RBM) can be stacked into a deep learning model, called deep belief network [45]. DBNs extracts hierarchical representation of the training data, as well as reconstruct their input data [43]. The efficiency of DBN as deep learning model comes from the fact that the training of a DBN is performed layer by layer, where each layer is treated as an RBM trained on top of the previous trained layer [45]. DNB is trained by a two-stage approach similar to the previously mentioned one for training SAEs. DBNs are suitable for hierarchical features discovery [43].

\subsubsection{Reinforcement learning}\mbox{} \\
In reinforcement-learning (RL), the system learns based on a set of reinforcements from its environment. For instance, a reward or punishment determines whether the system performed well or not [36]. Each interaction with the environment returns an information, which the system makes the best use of this information to learn and update its knowledge [59]. A key concept in reinforcement-learning is the Markovian property (i.e. only the current state affects the next state) [59].

\paragraph{Q-learning}\mbox{} \\
Q learning a form of model-free reinforcement learning allows agents to act optimally in controlled Markovian domains without the need for building maps of these domains [60]. The task for the agent is determining an optimal policy that maximizes the total discounted expected reward, called also Q, for executing a particular action at a particular state [60]. Q learning is classified as incremental dynamic programming because it finds the optimal policy in step-by-step manner [60]. Watkins [60] has proved that Q-learning converges with probability one under reasonable conditions on the learning rates and the Markovian environment. 

\paragraph{Deep reinforcement-learning}\mbox{} \\
Deep learning gives reinforcement-learning (RL) the ability to scale-up to decision-making problems with high dimensional state and action spaces [59]. Deep RL basically depends on deep neural networks for approximating the optimal policy [59]. Deep RL can leverage the representation learning to deal with the problem of the curse of dimensionality [59]. For example, it can use the convolutional neural networks (CNNs) to learn from high dimensional raw data [59]. AlphaGo represents one of the promising success for deep RL, which defeated the world champion in Go. AlphaGo depended on neural networks that were trained using supervised learning, RL and a traditional heuristic search algorithm [59].

\subsubsection{Semi-supervised learning}\mbox{} \\
In semi-supervised learning the system learns from both labelled and unlabelled data, where the lack of labels, as well as the labeled part may contain a random noise forms a situation between supervised and unsupervised learning [36]. It is more realistic as in many real-world applications, it is often difficult to collect many labelled data due to the fact that the data are manually labelled by the experts, whereas it more easier to collect a lot of unlabelled data [61]. As it includes some small labelled data, the performance of semi-supervise learning approaches is superior to unsupervised learning [61]. Table 1 shows a comparison between the main types of ML approaches.

\begin{table*}[ht!]
\centering
\caption{Comparison of ML approaches [62-64]}
\begin{tabular}{|c|c|c|}
\hline 
ML-approach & Advantages & Disadvantages \\ 
\hline 
\makecell{Supervised \\ Learning } & \makecell{Learns from labelled data. \\ Generalizes well based on a sufficient dataset.} & \makecell{Requires a dataset that represents the system.\\ The data is manually labelled by human experts, \\ which is not appropriate for many real-world applications.}
 \\ 
\hline 
\makecell{Unsupervised \\ Learning}& \makecell{Finds hidden patterns without relying on labelled data. \\ Performs better for unseen data in compared \\ with supervised approach.} & \makecell{
It may not provide a useful insight into \\ the hidden patterns and what actually they mean.} \\ 
\hline 
\makecell{Semi-supervised \\ Learning}& \makecell{Learns from both labelled and unlabelled data. \\ Certain assumptions about the underlying data \\ distribution must be met.} & \makecell{It may lead to worse performance when
\\ we choose wrong assumptions. } \\ 

\hline 
\makecell{Reinforcement \\ Learning}& \makecell{Dynamically adaptation and gradually refinement. \\ An agent interacts with an uncertain environment, \\ in which the goal is maximize the agent's reward. \\ It can also be used for difficult problems \\ that have no analytic formulation. } &  \makecell{ There is a trade-off between exploration and exploitation.\\
In addition, we need to specify a reward function, \\ parameterized policy, strategy and initial policy. }  \\ 
\hline 
\end{tabular} 
\end{table*}

\subsection{Meta-heuristic Algorithms}
The heuristic algorithms use a problem-specific heuristic, whereas the meta-heuristic algorithms form an efficient general purpose approach that includes a wide range of application areas ranging from finance to engineering and networking [65]. Meta-heuristics have been increasingly employed to solve hard optimization problems [66] that cannot be solved by any deterministic (exact) approach within a reasonable time [65]. Most of these algorithms are nature-inspired [66], from simulated annealing (SA) [67] to genetic algorithms (GA) [68], and from ant colony optimization (ACO) [69] to whale optimization algorithm (WOA) [70]. These algorithms are widely used to solve difficult problems such as combinatorial, highly non-linear and multi-modal optimization problems [71]. Each one of these algorithms has different advantages, therefore many efforts were made to design hybrid approaches that combine their benefits and ultimately attain better results [71]. The success of these algorithms is determined by achieving a balanced performance between the exploration and the exploitation [65]. Exploitation is useful for determining the most promising high-quality solutions in the search space, whereas exploitation is needed to concentrate search in some areas based on previous search results [65]. The main disadvantage of meta-heuristics is that these methods find good solutions rather than a guaranteed optimal solution. Another drawback is that these methods include a large number of parameters that need to be set in order to produce a good solution [72].

\subsubsection{Ant colony optimization (ACO)}\mbox{} \\
Ant colony optimization (ACO) [69] is a swarm intelligence population-based meta-heuristic algorithm for finding the solution of combinatorial optimization problems. ACO was inspired by the foraging behaviour of ants in nature. At first, the ants start randomly exploring the area surrounding their nest. Along the path they selected ants deposit a chemical pheromone trail, which guides the other ants to the food sources founded by the previous ants. After a period of time, the concentration of pheromone will increase along the shortest path of the food source. Pheromone evaporation helps in avoiding the problem of premature convergence [65].

\subsubsection{Evolutionary algorithms}\mbox{} \\
Evolutionary Algorithms (EA), also known as Evolutionary Computation (EC), is the main term for many optimization algorithms that are developed based on Darwinian theory of nature's capability to revolve and survival of the fittest [65]. EA domain includes genetic algorithms, evolution strategies, evolutionary programming, and genetic programming [65].

\paragraph{Genetic algorithm}\mbox{} \\
The Genetic Algorithm (GA) [68,73] is one of the well-known and mostly used population-based technique. In GA, a solution of an optimization problem is represented by a chromosome. A set of chromosomes forms the population. Two basic yet very important operations in GA are: crossover and mutation. The crossover operation combines previously selected individuals together by exchanging some of their parts. Mutation, on the other hand, brings some randomness into the search to avoid the problem of local optima. The important factors for implementing any GA are: the selection strategy and the type of crossover and mutation operators [65].

\subsubsection{Particle swarm optimization}\mbox{} \\
Particle swarm optimization (PSO) [74] is also another swarm intelligence, population-based, meta-heuristic algorithm that computationally mimics the flocking behaviour of birds to solve optimization problems. In PSO, a swarm consists of N particles, which are  stochastically generated in the search space. Each particle is represented by a velocity, a location, and has memory for remembering the best positions (solutions). PSO has succeeded at finding optimal regions of the search space. However, it has no feature that allows it to converge on optima. 

\subsubsection{Simulated annealing}\mbox{} \\
Simulated annealing (SA) [67] is a single-solution based meta-heuristic algorithm inspired by the annealing technique used to obtain a well ordered solid state of minimal energy. The objective function of a problem in SA is then minimized by introducing the temperature parameter T, which represents the main parameter of the algorithm [65]. At each iteration, SA selects a random solution from the neighbourhood of the current solution. The new solution is accepted based on the value of the objective function and the value of the parameter T, which decreases during the search process [65].

\subsubsection{Bee colony optimization-based algorithms}\mbox{} \\
Bee colony optimization-based algorithms are new swarm-intelligence-based algorithms motivated mainly by the collective behaviour of honeybee colony [65]. Artificial bee colony (ABC) [75] is one of the well-known and common used foraging-inspired optimization algorithm, which makes use of the bees' decentralized foraging behaviour. Interestingly, honey bees balance between exploiting known food sources and exploring potentially better food sources in the surrounding environment [65]. Bees in hive are divided into three sets: employed bees, onlooker bees and scouts. The number of employed bees is the same as the number of available food sources. When a food source is consumed an employed bee becomes a scout bee, which randomly searches for new food resources. Employed bees share the information about food resources with a certain probability using waggle dance [65]. ABC has a global search ability implemented through neighbourhood source production mechanism [76,77]. In addition to ABC, many other new algorithms have been developed based on the cooperative behaviour of social honey bees [77,80-82]. For more details, we refer the reader to [81].

\subsubsection{Whale optimization algorithm (WOA)}\mbox{} \\
Whale optimization algorithm (WOA) [70] computationally mimics the social behaviour of humpback whales when hunting their prey. 
Humpback whales has their special hunting method, called bubble-net feeding, which is done by creating distinctive bubbles along a circle or 9-shaped path. They have two manoeuvres associated with bubble namely upward-spirals and double-loops. They dive around 12m down and then they create bubble in a spiral shape around their prey and swim up toward the surface. Mirjalili and Lewis mathematically modelled this hunting behaviour in [70]. WOA has showed better results when compared with other meta-heuristic algorithms such as PSO [70].

\subsubsection{Firefly optimization (FFO)}\mbox{} \\
Firefly optimization (FFO) [82] is a population-based, meta-heuristic algorithm inspired by the social (flashing) and communication behaviour of fireflies. Brighter firefly attracts other fireflies (i.e. the brightness of a firefly indicates the goodness a solution). The attractiveness exponentially decreases based on the distance between them [83,84]. The main advantage of FFO is the fact that it uses real random numbers, and it depends on the global communication among the swarming fireflies [85]. FA, on the hand, performs a full pair-wise comparison, which might be considerably time consuming [84]. In addition to the fact that the light absorption coefficient is widely assumed to be equal to unity. However, as the distance between fireflies increases, fireflies might get trapped in their positions during the repeated evaluation of the algorithm [84].

\subsubsection{Bat algorithm (BA)}\mbox{} \\
Binary algorithm (BA) [86] is a new meta-heuristic method based on the fascinating capability of microbats to find their prey and distinguish between different types of insects even in complete darkness by decreasing the loudness and increasing the rate of emitted ultrasonic sound that bounces back from the surrounding objects [87,88]. An artificial bat has three vectors: position vector, velocity vector, and frequency vector. BA forms a balanced combination of PSO and intensive local search [87]. The balancing between these techniques is controlled by the parameters of loudness and pulse emission rate [87]. Yang [86] showed that the BA is able to outperform PSO and GA in terms of enhanced local optima avoidance and convergence speed. BA, however, could not be used to solve binary problems. Therefore, other researchers proposed a binary version of this algorithm called binary bat algorithm [87].

\subsubsection{Teaching-learning-based optimization (TLBO)}\mbox{} \\
TLBO [89] is a new population-based algorithm basically proposed for constrained mechanical design optimization problems. In TLBO, a set of learners is defined as population, and the optimal solution in the population is considered as the teacher [90]. TLBO needs merely the population size and the number of iterations as the control parameters [90]. Both the teacher and learner procedures are executed in each iteration of TLBO. When the teacher finds a better than the existing solution, it will be replaced with the new one [90].

\subsubsection{Grey wolf optimization (GWO)}\mbox{} \\
GWO is a population-based algorithm that computationally mimics the social hierarchy and hunting behaviour of Grey Wolves [91].
The first level of leaders, called alphas, responsible for making decisions about different actions that should be followed by the pack. The first level in the hierarchy of grey wolves is alphas followed by beta, delta and omega respectively [91]. When gray wolves find a prey, they encircle it and then they start attacking the pray. Mirjalali et al. modelled this hunting behaviour in [91].

\subsection{Fuzzy Inference Systems}
Unlike classical binary logic where any fact can only be true or false, fuzzy logic (FL) is a multi-valued logic, which deals with a degree of truth or degree of membership (i.e., any value between 0 and 1) [37]. Therefore, Boolean logic can be seen as a special case of FL. Fuzzy systems (FS) are used for mapping a given input to an appropriate output based on the principles of fuzzy set theory, which was developed by Lotfi Zaheh [37, 92]. FS make use of fuzzy rules in the form:

\begin{equation}
\label{eq2}
If \ x \ is \ A \ and \ y \ is \ B \ \ then \ \ z \ is \ k
\end{equation}
where x, y and z are linguistic variables; A and B are linguistic values determined by fuzzy sets on the universe of discourses X, Y respectively and k is a constant that represents the consequent of the rule. As shown in Fig. 3, the system converts the crisp input to the appropriate fuzzy sets using the membership functions. Thereafter, it will be evaluated using the inference engine. Finally, the output is determined using an appropriate defuzzification method such as center of gravity (COG) or weighted average (WA) [37].

\begin{figure}
\centering
\includegraphics[width=80mm]{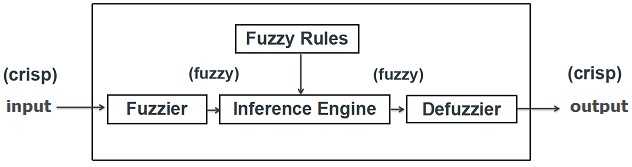}
\caption{A structure of a typical fuzzy inference system\label{overflow}}
\end{figure}

Mamdani and Sugeno methods are two inference techniques that can be used in fuzzy systems. Unlike Mamdani's approach, which employs a fuzzy membership function of the rule consequent, Sugeno's approach uses a single point to represent the rule consequent. Therefore, Sugeno's approach, is considered to be more efficient than Mamdani's approach which requires finding the centroid of a two-dimensional shape and this, in turn, increases the computational cost [37]. 

The main advantage of using fuzzy systems is the human-like knowledge representation and their explanation abilities. However, fuzzy systems cannot learn or adjust themselves to a new environment. In addition, building these models often requires tuning fuzzy sets and fuzzy rules to meet certain requirements. Hybird methods such as neuro-fuzzy systems combine the learning abilities of neural networks with the representation and explanation abilities of fuzzy systems [37].

\section{Artificial Intelligence in SDN}
AI and ML approaches have been widely used for solving various problems such as routing [93], traffic classification [94], flow clustering [95] intrusion detection [96], load balancing [97], fault detection [98], quality of service (QoS) and quality of experience (QoE) optimization [99], admission control and resource allocation [100]. However, in the era of SDN the role of AI was significantly increased due to the huge efforts made by industry and research community. Recent studies have shown a strong tendency of research community towards adoption of AI approaches in SDNs. It is worth mentioning that ML, meta-heuristics and fuzzy systems were the most common approaches for solving various networking related problems. We investigate the application areas of each approach in the SDN paradigm separately as will be shown in the reset of this study.

\subsection{Machine Learning in SDN}
\subsubsection{Supervised learning in SDN}\mbox{} \\
In this context, neural networks [101,103-106,108-112], support vector machine [111-119], decision trees [120-134], ensemble methods [135-146] and supervised deep learning [44,149-153] approaches were the most used supervised learning techniques in SDN.

\paragraph{Neural Networks in SDN}\mbox{} \\									
Neural networks (NN) approach was used mainly for intrusion detection and prevention [101,106,112], solving controller placement problem [104], load balancing [106], performance prediction [108], service level agreements (SLA) enforcement [109,110], routing and optimal virtual machine (VM) placement [111].

Chen and Yu [101] proposed, CIPA, a collaborative intrusion prevention architecture, which represents a distributed intrusion prevention system based on NN approach. They used the following features: number of all packets monitored, proportion of icmp packets to all packets, the proportion of short packets, the proportion of long packets, : the proportion of udp packets to all packets and : the base-10 logarithm of the proportion of packets with syn flag set to packets with ack flag set. The experimental results showed that CIPA outperforms [102] in detecting DDoS flooding attacks. CIPA also achieved good results in detecting Witty, Slammer and Conficker worm outbreak. The system achieved low computational and communication overhead due to its parallel and simple computational capabilities. 

He et al. [103] proposed a multi-label classification approach to predict global network allocations (i.e. weighted controller placement problem). Compared to decision tree and logistic regression, neural network approach showed better results and saved up to two-thirds of the algorithm runtime. Alvizu et al. [104] used a neural network approach for off-line prediction of traffic demands in a mobile network operator, which resulted in reducing the optimality gap below 0.2\% (virtual wavelength path-hourly-NN) and 0.45\% (wavelength path-hourly-NN). In addition, NN approach was used for off-line prediction of the next configuration time point. Abubakar et al. [105] proposed an intrusion detection system for SDN based on NN approach, which achieved a high accuracy reached 97.3\% using NSL-KDD dataset.                                                                                                                                                                                              

Chen-Xiao and Ya-Bin [106] proposed an NN approach for load balancing. Compared to [107] and static Round Robin strategy, the  experimental study showed that this method can achieve better performance and resulted in 19.3\% decreasing of network latency.                                                      Sabbeh et al. [108] proposed an NN approach based on Levenberg Marquardt (LM) algorithm to predict the performance of SDN according to round-trip time (RTT) and throughput which are the two parameters used for training. The experimental results showed that one hidden layer achieves low mean squared error (MSE).

Bendriss et al. [109,110] proposed a new approach to enforce service level agreements (SLA) in SDN and virtualized network functions (NFV). Their research focused on prediction of service level objectives (SLOs) breaches for streaming services running on  NFV and SDN. The experimental results showed that long short term memory (LSTM) is more robust than feed forward neural networks.

Mestres et al. [111] presented a new paradigm that employs AI in SDN, termed as knowledge-defined networking (KDN). The paradigm included a proof of concept concerning routing in an overlay network based on NN approach where the mean squared error (MSE) reached 1\%. In addition to solving the problem of optimal virtual machine (VM) placement in NFV paradigm. Mihai-Gabriel and Victor-Valeriu [112] proposed a method for mitigating DDoS attacks in SDNs by risk assessing based on neural networks and biological danger theory. The risk of a DDoS attack is calculated on every host and then the reported to a VM that is responsible for monitoring the network traffic. When the risk of the observed traffic exceeds a predefined value, instructions will be sent to the controller to install the appropriate controls that allow the SDN to enter in a proactive mode to reduce the burden on the controller caused by sending these flows to the controller for analysis.

\paragraph{Support Vector Machines in SDN}\mbox{} \\
Support vector machine (SVM) approach was mainly used for deploying intrusion detection systems in the SDN paradigm [113-120]. Kokila et al. [113] proposed a method for detection of DDoS attacks on the SDN controller. SVM showed higher accuracy and less false positive rate when compared to other classifiers. Phan et al. [114] proposed, OpenFlowSIA, a framework for detection of flooding attacks based on SVM and idle-timeout adjustment (IA) algorithm, which leads to an accurate system and reduction  of CPU usage of Open vSwitches and their correspondence SDN controller. 

On the other hand, an SVM approach for network intrusion introduced in [115]. The model also employed the ID3 decision tree approach for feature selection. Based on the experimental results conducted on KDD CUP99 dataset, the system showed an accuracy of 97.60\%. Boero et al. [116] used SVM for SDN-based malware detection where information gain (IG) metric was used for selecting the most relevant features. Their model achieved a detection rate of 80\% and 95\% for malware and normal traffic respectively. In addition, it showed a false positive rate of 5.4\% and 18.5\% for malware and normal traffic respectively.

Phan et al. [117] proposed a new approach that combines SVM with self-organizing map (SOM) in order to get a higher accuracy and better detection rate for DDoS attacks in SDN, as well as achieving lower false alarm rate.  The input vector for the SOM module is a 4-tuple including four attributes: number of packet, number of byte, duration and protocol. The experimental results showed that this system was able to achieve an accuracy of 97.6\% and a false positive rate of 3.85\%. Shang et al. [118] proposed an SVM approach for implementing a traffic-based filtering classifier, which represents the second stage of a framework proposed to mitigate DoS attacks. Their model showed a high detection rate when attack rate is higher than 3000 packets per second.                                       

Hu et al. [119] proposed ,FADM, a real-time lightweight framework for detecting and mitigating DDoS attacks in SDN. The real-time attack detection module works as an application in the controller. The attack mitigation module consists of two main components: 1) Mitigation server and 2) Mitigation agent. The mitigation server works as an application in the controller whereas the mitigation agent runs on a host in the SDN network. In the detection stage, the authors collected the following features: source IP address entropy, destination IP address entropy, source port entropy, destination port entropy and protocol type. The authors proposed to use controller-based method and sFlow-based method according to different network environments. The controller based method (i.e. based on the incoming Packet-In messages) is efficient for low rate attacks. Whereas sFlow-based method (i.e. flow samples generated periodically from sFlow agents embedded in OF-enabled switches) is efficient for high rate attacks. FADM was implemented on the POX controller. TFN2K tool was used to launch multiple types of DDoS flooding attacks. The SVM algorithm was trained on a relatively small training dataset which consists of 552 attack samples and 662 benign samples. The attack mitigation mechanism was based on white-list and dynamic updating of forwarding rules. Compared to the controller-based method, the sFlow-based method showed higher detection rate reached 100\% when the attack rate was greater than 3000 packets per second (pps). In terms of the average detection, the difference between these methods is very small.

Latah and Toker [120] introduced a two-stage SDN-based approach for detecting DoS attacks. The detection loop consists of the following two stages: 1) Calculation of packet rate on the controller side and 2) SVM classification on the host side. When the packet rate exceeds a predefined probability, then the system will activate a host-based packet inspection unit, which collects packet-based statistics during a 4-second period and then uses the RBF-SVM algorithm in order to determine whether the attack is happened or not. The authors generated a dataset which consists of 321 instances of normal traffic and 639 instances of DoS flooding attack. Thereafter, the system was evaluated using 10-fold cross-validation and achieved 96.25\% accuracy with 0.26\% false alarm rate.

Rego et al. [121] presented an intelligence system to detect problems and correct errors in multimedia transmission in surveillance SDN-based IoT environments. Their proposed AI module consists of two different parts. The first one is the traffic classification part based on SVM algorithm, which detects the type of network traffic. The second part is an estimator that informs the SDN controller on which kind of action should be taken to guarantee the quality of service (QoS) and quality of experience (QoE). The experimental results showed that the jitter was reduced up to 70\% on average and losses were reduced from 9.07\% to nearly 1.16\%. Furthermore, SVM was able to detect critical traffic with an accuracy of 77\%.                                              

Bouacida et al. [122] employed supervised approaches to detect long-term load on SDN controllers. They generated a dataset contains nine features and 2344 instances (799 instances labeled as long-term load, 1545 instances labeled as short-term load) based on injecting Packet-In messages to the controller. The experimental results showed that the linear SVM approach achieved the best results among k-nn and Naive Bayes approaches in terms of accuracy, precision, F1-measure and Area Under AUC curve. In addition, the real-time evaluation showed that both of accuracy and precision values diminish proportionally when they use larger look-ahead intervals. Moreover, offloading delay increases linearly with the number of flow entries.

\paragraph{Decision Trees in SDN}\mbox{} \\
Decision trees were widely used for application identification [123, 124], packet and traffic classification [129,132], intrusion detection [125,135,136], Botnet detection [128,136], solving flow table congestion [130], detection of elephant flows [133], prediction of QoS violations [134], as well as solving SDN-related security challenged [126,131].  

In [123] and [124] the C4.5 decision tree approach was used for application identification in order to associate each application type with a corresponding QoS level. Le et al. [125], on the other hand, proposed an intrusion detection and prevention system based on C4.5 algorithm. Nagarathna and Shalinie [126] proposed, SLAMHHA, a supervised learning approach based on Iterative Dichotomiser 3 (ID3) decision tree algorithm for mitigating host location hijacking attacks on SDN controllers.                                                                                                                                                                                          This attack is a new attack vector for SDNs where it exploits the vulnerability of the SDN to launch a DoS attack on the controller. By exploiting the vulnerability of the host tracking service of the SDN controller, which estimates the host location by examining the PACKET-IN messages. SLAMHHA consists of three main components: 1) decision tree construction, 2) modified host tracking service module and 3) classification module. When a new PACKET-IN message is received, SLAMHHA probes the message to investigate the host related information. A check is made to find whether the host related information is presented in the host profile. If no match is found, then the classification event is invoked by the SDN controller. SLAMHHA achieved less overhead in terms of CPU and memory consumption when compared to the authentication method.

Tariq and Baig [127] applied C4.5 for SDN-based Botnet detection. In addition to the OpenFlow statistics they extracted four more features namely: average packet inter-arrival time , bytes per packets, bits per second, and packets per second. The training phase included four Botnets whereas the testing phase included five additional Botnets to check the detection capability of the proposed model. The experimental results based on CTU-43 botnet dataset showed that proposed approach achieved an accuracy of 80\%. 

Qazi et al. proposed [128], Atlas, a framework that enables fine-grained and scalable application classification for mobile agents in SDN based on C5.0 algorithm. Atlas was prototyped in HP Lab wireless network and it collected the netstat logs using mobile agents running on employee devices. These logs are sent to the control plane, at which the classification algorithm works. The flow features were collected by an OpenFlow enabled wireless AP and sent to the control plane. The authors used 40 most popular applications in Google Play Store in order to collect over 100k flow samples from five devices during 3 weeks of the testing period. The experimental results showed that Atlas can achieve a 94\% of accuracy on average.

Leng et al. [129] presented a method for solving flow table congestion problem based on C4.5 algorithm.                                                                                                                                                                                                                                                                             
C4.5 is employed to compress the flow entries with QoS guaranteed. DT is translated into a new flow table according to the path between each node and the root. The experimental results showed that the proposed approach achieved a higher compressing with larger number of flow entries and also it reduced the flow matching cost (average matching time dropped by 98\% on average). Nanda et al. [130] proposed a method for predicting potential vulnerable hosts using historical network data and ML algorithms such as C4.5, bayesian network (BN), decision table (DT), and naive-bayes (NB). The best results were achieved by bayesian network approach.                                                           

Stimpfling et al. [131] introduced new extensions for decision-tree algorithms, namely adaptive grouping factor (AGF) and independent sub-rule (ISR) leaf structure that allow achieving better packet classification for larger rules, as well as reducing the number of memory access by a factor of 3. Furthermore, it decreased the size of the data structure by about 45\% over EffiCuts approach. The authors considered SDN rule-sets for this study.

Tang et al. [132] proposed a two-stage approach for detecting elephant flows, where the first stage consists of an efficient sampling used in order to attain a good balance between detection overhead and implementation. The the next stage, on the other hand, employs an enhanced C4.5 for the previously correlated flows. The proposed approach depends mainly on finding the most effective sampling periods and classifying these samples based on enhanced C4.5. The new C4.5 is based on flow correlation and probability. It is worth mentioning that the proposed system requires only one packet to identify whether a particular flow is an elephant flow or not. The enhanced C4.5 improved the accuracy of C4.5 up to 12\%.

Jain et al. [133] investigated the prediction of traffic congestion based on M5Rules approach, which combines decision trees and linear regression in order to improve the management of QoS. They proposed a multi-dimensional analysis of Key Performance Indicators (KPIs) followed by M5Rules decision tree to discover different types of correlations, which have been divided into 3 groups: expected correlations, discovered correlations and unexpected correlations. Van et al. [134] used a J48-tree classifier for intrusion detection on OpenFlow switches. They implemented an FPGA-based prototype where the experimental results based on  KDD99 dataset showed that their proposed system can achieve an overall accuracy of 93.3\% and a detection rate of 91.81\% with low false alarm rates (0.55\%). Wijesinghe et al. [135] studied the detection of Botnets (IRC,HTTP and P2P botnets) using SDN paradigm based decision trees approach, which showed better results for detecting peer-to-peer botnets, whereas SVM and Bayesian networks were more effective in detecting command and control (C\&C) related Botnets such as HTTP and IRC (internet relay chat) Botnets. 

Latah and Toker [136] presented a comparative analysis of application of different supervised ML approaches for SDN-based intrusion detection task. They used Principal Components Analysis (PCA) for feature reduction. The experimental results based on NSL-KDD dataset showed that DT approach achieved the best performance in terms of accuracy, precision, F1-measure, AUC and Mc Nemar's Test. Whereas Bagging and boosting approaches achieved better results over traditional supervised ML approaches such as k-NN, NN and SVM. LogitBoost also showed the best results in terms of false alarm rate and recall.
       
\begin{table*}[ht!]
\scriptsize
\addtolength{\tabcolsep}{-2pt}
\centering
\caption{Conventional supervised ML approaches in SDN paradigm}
\hskip-1cm
\begin{tabular}{|c|c|c|c|}
\hline 
Reference & \makecell{ Supervised ML approach } & Task &  Findings \\ 
\hline 
Chen and Yu [101] & NN & Collaborative intrusion prevention & \makecell{Outperformed [102]. Also it achieved a low overhead \\ due to its parallel and simple computational capabilities.}  \\ 
\hline 
He et al. [103] & NN & Solving weighted controller placement problem & Outperformed decision tree and logistic regression.  \\ 
\hline 
Alvizu et al. [104] & NN & \makecell{Off-line prediction of traffic demands \\ in a mobile network operator} & \makecell{ Reduced the optimality gap below 0.2\% (virtual wavelength path-hourly)\\ and 0.45\% (wavelength path-hourly). }  \\ 
\hline 
Abubakar et al. [105] & NN & Intrusion detection & \makecell{ An accuracy of 97.3\% using NSL-KDD dataset.}  \\ 
\hline 
\makecell{Chen-Xiao and \\Ya-Bin [106]} & NN & Load balancing & \makecell{ Compared to [107] and static Round Robin strategy, NN achieved \\ better performance and 19.3\% decreasing of network latency. }  \\ 
\hline 

Sabbeh et al. [108] & NN & Predicting the performance of SDN & \makecell{Achieved low mean squared error (MSE).}  \\ 
\hline 

Bendriss et al. [109,110] & NN & SLA enforcement in SDN and NFV & \makecell{Showed less robust in compared with LSTM.}  \\ 
\hline 

Mestres et al. [111] & NN & Routing in an overlay network & \makecell{Mean squared error(MSE) reached 1\%.}  \\ 
\hline 

\makecell{Mihai-Gabriel \\ and Victor-Valeriu [112]} & NN + biological danger theory & Mitigating DDoS attacks in SDNs & \makecell{ Proposal without simulated proof of applicability. }  \\ 
\hline

Kokila et al. [113] & RBF-SVM & DDoS attack detection & \makecell{ An accuracy of 95.11\% and false positive rate of 0.01\%.}  \\ 
\hline 

Phan et al. [114] & Multiple Linear SVM & DDoS attack detection & \makecell{Reduction of the consumption of SDN's resources.}  \\ 
\hline 
Wang et al. [115] & RBF-SVM & DDoS attack detection & \makecell{An accuracy of 97.60\%. } \\ 
\hline 
Boero et al. [116] & RBF-SVM & \makecell{ Malware Detection} &
\makecell{ A detection rate of 80\% for malware 95\% for normal traffic.\\ False positive rate of 5.4\% for malware 18.5\% for normal traffic. }
\\ 
\hline 
Phan et al. [117] & \makecell{ Multiple Linear\\ SVM + SOM } & DDoS attack detection &
 \makecell{An accuracy of 97.6\% and false positive rate of 3.85\%. } \\ 
\hline 
\makecell{ FloodDefender \\ Shang et al. [118] } & SVM & DoS attack detection & \makecell{ Attack detection rate of 96\% with less \\  than 5\% of false-positive rate.}  \\ 
\hline 
\makecell{ FADM \\ Hu et al. [119]} & SVM & DDoS attack detection & \makecell{ High detection rate when
attack rate is higher than\\ 3000 packets per second.}  \\ 
\hline 
Latah and Toker [120] & RBF-SVM & DoS attack detection & \makecell{ An accuracy of 96.25\% with false positive rate of 0.26\%.} \\ 
\hline 
Rego et al. [121] & SVM & Traffic classification & \makecell{SVM was able to detect critical traffic with an accuracy of 77\%.} 
\\ 
\hline 
Bouacida et al. [122] & Linear-SVM & Detecting long-term load on SDNs & \makecell{SVM outperformed k-NN and Naive Bayes. } \\ 
\hline 
Li et al. [123] & C4.5 & Application identification & An average accuracy of 99\%. \\ 
\hline 
Pasca et al. [124] & C4.5 & Application identification & \makecell{ An accuracy of 98\%. outperformed
Naive Bayes, Naive Bayes Kernel \\ Estimation, Bayesian Network and SVM.} \\ 
\hline 
Le et al. [125] & C4.5 & Intrusion detection and prevention & \makecell{High precision, recall with low false positive rate.} \\ 
\hline 
\makecell{Nagarathna and \\ Shalinie [126]}  & ID3 & \makecell{Mitigating host location hijacking \\ attacks on SDN controllers} & 
\makecell{Less overhead in terms of CPI and memory \\ consumption compared to authentication method.}  \\ 
\hline 
Tariq and Baig [127] & C4.5 & Botnet detection & An accuracy of 80\%. \\ 
\hline 
Qazi et al. [128] & C5.0 & \makecell{Fine-grained and scalable\\ application classification} & An average accuracy of 94\%. \\ 
\hline 
Leng et al. [129] & C4.5 & \makecell{Solving the problem of flow table congestion} & \makecell{High compression with large  number of flow entries \\ and reduced the flow matching cost.} \\ 
\hline   
Nanda et al. [130] & C4.5 & \makecell{Prediction of potential vulnerable hosts} & \makecell{Outperformed NB and decision table. The best results, \\however,achieved by bayesian network.} \\ 
\hline 
Stimpfling et al. [131] & \makecell{ Extensions \\for DTs} & \makecell{New extensions for DTs for better packet \\ classification and lower memory access } & \makecell{Better packet classification for larger rules, reducing the number \\ of memory access by a factor of 3 , and decreasing the size of \\data structure 45\% over EffiCuts.} \\ 
\hline 
Tang et al. [132] &  \makecell{ Enhanced C4.5} & Detection of elephant flows & \makecell{Improve the accuracy of C4.5 up to 12\%, \\ recall rate 88.3\%, false positive rate less than 2.13\%.} \\ 
\hline 
Jain et al. [133] & M5Rules & Prediction of QoS violations & Discover different types of correlations.  \\ 
\hline 
Van et al. [134] & J48-tree & \makecell{Intrusion detection on OF switches} & 
\makecell{An overall accuracy of 93.3\% and detection rate \\ of 91.81\% with low false alarm rates 0.55\%.} \\ 
\hline 
Wijesinghe et al. [135] & DT & Botnet detection & \makecell{DT showed better results for detecting P2P Botnets whereas SVM and \\Bayesian networks showed effectiveness in detecting C\&C Botnets.}  \\ 
\hline 
Latah and Toker [136] & \makecell{Comparing different\\supervised ML algorithms} & SDN-based intrusion detection & \makecell{DT achieved the best level of accuracy over other supervised ML approaches. \\ However, ensemble methods achieved the best false positive rate.} \\ 
\hline 
Stadler et al. [137] & RF & Estimating service-level metrics & \makecell{Outperformed regression tree (RT) in terms of estimation accuracy. \\ However, RF is 3x longer than RT in terms computation time.} \\ 
\hline 
Song et al. [138] & RF & Intrusion detection & An accuracy of 0.99\% on KDD99 \\ 
\hline 
Miettinen et al. [139] & RF & \makecell{Automatic identification and \\ security enforcement for IoT devices} & \makecell{ An accuracy of 0.815\% and low execution time (<1 ms).} \\ 
\hline 
Abar et al. [140] & RF & QoE prediction & Outperformed k-NN, NN and DT. \\
\hline 
Amaral et al. [141] & \makecell{RF} & Traffic classification & \makecell{RF achieved competitive results with the stochastic \\  gradient boosting and extreme.} \\ 
\hline  
Zago et al. [142] & RF & Cyber threat detection & \makecell{Outperformed k-NN, naive bayes and logistic regression.} \\ 
\hline 
Ajaeiya et al. [143] & RF & Cyber threat detection & \makecell{Outperformed k-NN, naive bayes, bagged-trees \\and logistic regression in terms of F1-score.}\\ 
\hline 
Anand et al. [144] & RF & Detecting compromised controller & \makecell{Outperformed Naive Bayes, SVM, MLP and AdaBoost.} \\ 
\hline 
Hussein et al. [145] & RF & Intrusion detection & \makecell{Outperformed SVM, k-NN, DT, NN and DNN.} \\ 
\hline 
Su et al. [146] & RF & Botnet detection & \makecell{An average accuracy of 99.77\% } \\ 
\hline 
Chen et al. [147] & XGBOS & DDoS attack detection & \makecell{ Outperformed RF, GBDT and SVM in \\ terms of accuracy and false positive rate.} \\ 
\hline 
Choudhury et al. [148] & \makecell{RF and Gradient boosted \\ regression trees. } & \makecell{ Prediction of traffic matrix and \\ performance of optical path.} & \makecell{Outperformed rigde regression, LASSO regression, LASSO with quadratic\\ features,  MLP, Guassian process regression, gradiant boosted regression trees.} \\ 
\hline 

\end{tabular}
\end{table*}

\paragraph{Ensemble methods in SDN}\mbox{} \\
The random forest (RF) approach was widely used for estimating service-level metrics [137], intrusion detection, security applications [139,142,143-147], QoE prediction [140], traffic classification [141] and prediction of traffic matrix and performance of optical path [148]. 

Stadler et al. [137] proposed a RF approach for estimating service-level metrics based on network-level statistics. Song et al. [138] presented an intrusion detection system where the RF approach has been used for both feature selection and classification. Miettinen et al. [139] introduced, IoT SENTINEL, an SDN-based system for automatic identification and security enforcement of potentially vulnerable IoT devices using device type-specific profiling approach.                                                                   They used SDN for isolation and traffic filtering where the system can control the traffic flows of vulnerable devices to protect the other devices in the network from any potential threats. The device identification was achieved by fixed length fingerprints, where RF approach was used as a classification algorithm.

Abar et al. [140] proposed a ML based QoE prediction approach in SDN paradigm. The authors utilized the following ML algorithms: RF, k-nearest neighbours (k-NN), NN and decision trees where RF showed better performance. Amaral et al. [141] used several ML techniques such as RF, stochastic gradient boosting and extreme gradient boosting for traffic classification task. RF approach showed competitive results in compared with the other two algorithms in terms of the classifier accuracy. Zago et al. [142] proposed an RF approach for for cyber threat detection. RF approach showed the best performance among k-NN, naive bayes and logistic regression (LR) for cyber threat detection. Ajaeiya et al. [143] proposed a machine learning SDN-based approach for analysing, detecting and reacting against cyber threats. Experimental resulted based on ISOT botnet dataset showed that RF approach achieves the best results in terms of F1-score among k-NN, naive bayes, bagged-trees and logistic regression. 

Anand et al. [144] introduced a method detecting compromised controllers in SDNs. The authors identified five threat models for representing the compromised controllers. They used 9 OpenFlow-specific features in order to correctly and accurately build their ML model. The RF approach showed the highest accuracy (97\%) compared to Naive Bayes, SVM, NN, AdaBoost and DT. The proposed approach, however, was not able to exactly identify and locate the compromised controller when multiple physical controllers are included.

Hussein et al. [145] designed two architectures for building a general solution to defend and enhance the security of communication networks. The first architecture is distributed extraction, centralized processing, and centralized management.
The second one is distributed extraction, distributed processing and centralized management. Then the authors introduced a two-stage detection technique. The first stage includes detecting whether an attack happened or not, whereas the second stage includes identifying the type of attack. The experimental results conducted on the NSL-KDD dataset showed that the RF approach achieves better results when compared with the other techniques including SVM, kNN, DT, neural networks and deep learning.

Su et al. [146] presented an intelligent approach to detect P2P Botnets. The detection model consisted of two sub-modules namely:  primary classification module and secondary classification module. The primary classification module included one binary classifier for each application or Botnet. Whereas the secondary classification module employed a multi-class classification, which is used when more than one binary classifier matches in the primary classification module. The authors tested the performance of k-NN and SVM for the primary classification module. Whereas RF was used for the secondary classification module. Both of these stage were evaluated by 10-fold cross validation. In addition, they collected network traffic samples generated by different P2P Botnets and normal P2P applications. They used traffic trace files of Storm and Zeus Botnet as malicious training samples of P2P Botnets. Whereas network traffic trace files of eMule, uTorrent, and Skype, were used for as benign training samples of P2P applications. The following features were also selected: packet count, packet size, flow size, inter-arrival times (min,max, mean, and standard deviation), TCP Push flag count, duration, total bytes, TCP Urgent flag count. The experimental results showed that SVM achieved better results compared to kNN. However, the authors used kNN for the primary classification, because the training time of SVM is much longer compared to kNN. In addition, RF was able to achieve an average accuracy of 99.77\% for the second stage.

Chen et al. [147] used XGBoost classifier for DDoS attack detection in SDN-based cloud. XGBoost is an enhanced version of the traditional Gradient Boosting Decision Tree (GBDT) and provides a more flexible approach for preventing fitting and increasing the generalization ability of the model. KDD Cup 99 dataset was used for training the model based on 9 important features, which have the maximum information gain and chi-square statistic. They used Mininet to simulate real DDoS attacks by an attack tool called Hyenae. XGBOS showed an accuracy of 98.53\%. Whereas RF, GBDT and SVM showed an accuracy of 96.33\%, 97.69\% and 97.19\% respectively. In addition, the false positive rate of XGBoost was 0.008 whereas RF, GBDT and SVM showed a the false positive rate of 0.018\%, 0.013\% and 0.011\% respectively. In terms of training time, however, the best results achieved by RF and the worst results achieved by SVM. 

Choudhury et al. [148] introduced two applications of ML approaches for managing IP and optical networks. The first one was prediction of network traffic matrix, which allows proactive network updates for providing more flexible services. The second one is prediction of optical path performance in multi-vendor network based on the latest optical performance data. For the first task, they employed Gaussian process regression (GPR). For the second task they employed machine learning methods that broadly fall into three categories: penalized linear regressions, non-linear regressions and ensembles of regression trees. Among ensemble models, they applied gradient boosted regression trees and random forests. RF achieved the best overall error rates. Research efforts made to apply conventional ML approaches in SDN are briefly summarized in Table 2.

\paragraph{Supervised deep learning in SDN}\mbox{} \\
Deep learning techniques have shown promising results in SDN compared to traditional ML approaches. Tang et al. [149] proposed an intrusion detection system based on a simple deep neural network, where deep learning approach achieved the best results compared to other supervised ML algorithms such as naive bayes, SVM and DT, with an accuracy of 75.75\%.                                      In [44], The same authors improved their results by using GRU-RNN instead of the simple deep learning model. Their new proposed model outperformed their previous simple DNN-based model [149], SVM and NB Tree with an accuracy of 89\%. Lazaris and Prasanna [150] proposed, DeepFlow, an intelligent traffic measurement framework for SDNs that uses the available TCAM memory to install measurement rules for important flows, and employs LSTM-RNN to predict the size of rest of the flows when flow counters cannot be placed at a switch due to its limited resources, on the basis of historical data from previous measurement periods. The experimental results showed an average mean absolute percentage error (MAPE) of 12\% for approximating real flow sizes on CAIDA traces and MAPE of 3.9\% on simulating the network topology of Google's B4.

Azzouni et al. [151] employed a (LSTM-RNN) approach for predicting traffic matrix. GÉANT traffic matrices and related network states were the input of the LSTM network. The authors implemented a GÉANT network topology and generated 10000 samples using Mininet. Experimental studies showed that this approach outperformed an efficient dynamic routing heuristic by finding the near optimal path in shorter time. In the same context, Azzouni and Pujolle [152] used (LSTM-RNN) for the same purpose. However, in this work they validate their model based on real-world data from GÉANT backbone networks. The experimental results showed that RRN-LSTM can outperform linear forecasting models (ARMA, ARAR, HW) and feed forward neural network (FFNN).

Huang et al. [153] studied adversarial attacks on SDN-based deep learning port scan detection system. They applied three different adversarial attack mechanisms namely: Fast Gradient Sign Method (FGSM), Jacobian-based Saliency Map Attack (JSMA), JSMA reverse (JSMA-RE) to three different deep learning algorithms: MLP (which is previously refereed as DNN in this paper), CNN, LSTM. The deep learning models detected the port scan attacks based on Packet-In messages and STATs reports. They generated samples with three different adversarial test sets and one normal test set. JSMA showed a significant effect on the deep learning models ranges from 14 to 42\%. JSMA-RE did not reduce the accuracy of CNN and LSTM, however it reduced the accuracy of MLP to 35\%. FGSM caused a significant reduction in the accuracy of LSTM (more than 50\%).

\subsubsection{Unsupervised learning in SDN}\mbox{} \\
K-means clustering [154-158], self-organizing map [159-164], hidden Markov model [54,165], Restricted Boltzmann machines [166], and unsupervised deep learning approaches [55,56-58] were the most used unsupervised learning techniques in SDN paradigm.

\paragraph{K-means clustering in SDN}
\mbox{}\\
\ Ivannikova et al. [154] proposed a method for detecting application layer DDoS attacks in SDN-based cloud environments based on k-means and probabilistic transition. They extracted the following features at every time interval: 1) duration of the conversation, 2) number of packets sent in 1 second, 3) number of bytes sent in 1 seconds, 4) average packet size, and 5) presence of packets with different TCP flags. First, clustering is used to divide the features into different groups that represent specific classes of network traffic. Then conversations with the same source IP, destination IP and destination port at a certain time interval are grouped together. Each session in every time window is represented by a sequence of cluster labels obtained from the first step. Finally, they estimate conditional and marginal probabilities from the previously obtained sequences. Then they compare it against a corresponding threshold value, if it is lower then it is marked as anomalous. The experimental results showed that the combination of k-means and the probabilistic transition approach achieves the best results when compared with both Clustering Using REpresentatives (CURE) with the probabilistic transition, and n-gram with k-means respectively.

Nguyen et al. [155] introduced a k-means approach for achieving Wi-Fi direct clustering in campus networks. Their simulation results showed that the proposed approach achieves better performances in terms of download time and packet error rate compared to the Wi-Fi infrastructure mode. Sahoo et al. [156] investigated the problem of optimal controller placement based on k-means approach. They considered two clustering algorithms namely: k-medoids and k-center. The experimental results showed that k-center algorithm achieves better result than K-Medoids. Bakhshi and Ghita [157] used k-means for user traffic profiling in campus SDN based on their application trend. They derived six unique user traffic profiles obtained from OpenFlow statistics, which were generated by realistic campus switch during two-week. Their proposed system showed minimum computational cost and low OpenFlow control overhead. Barki et al. [158] used Naive bayes, k-NN, k-means and k-medoids for detecting DDoS attacks in SDN, where the highest detection rate achieved by naive bayes approach. K-means, on the other hand, achieved better results over kNN and Naive Bayes in terms of processing time.

\paragraph{Self-Organizing Maps (SOM) in SDN}\mbox{} \\
Self-organizing map (SOM) approach was used widely in the SDN paradigm for intrusion detection [159-164]. Jankowski and Amanowicz [159] presented an intrusion detection based on SOM approach. Kohonen algorithm was used for training the SOM where the neuron whose weights are most similar to the input vector and its neighbours will update their weights. The classification step includes determining which neuron is activated under that particular input. The main disadvantage in this approach is high-grained traffic matching in flow tables. Wang and Chen [160] proposed, SGuard, a lightweight DoS attacks detecting and mitigating framework based on a combination of access control and SOM classification. They collected 6 traffic features: percentage of flows with a small number of packets, percentage of flows with small average bytes, percentage of flows with short time duration, percentage of reversible flows, growth rate of irreversible flows and growth rate of ports. Then, they proposed two novel feature ranking and feature selecting algorithms, which allowed the SOM to achieved high detection rate with less number of features. In addition, their proposed framework can deal with IP/MAC spoofing attacks.

Phan et al. [161] proposed, DSOM , a distributed SOM approach for tackling the performance bottleneck and overload problems for large-sized SDNs under flooding attacks. They developed an application, called DSOMController, that manages the operation of DSOM. At each switch, DSOM was trained using a dataset obtained from the DSOMController. The DSOM Controller collects the results, which are SOM maps, from OpenFlow switches to construct a final merged SOM map. DSOM used the following 6 traffic features: number of flows, number of packet per flow, number of bytes per flow, duration, growth of client ports, protocol type. They used k-means clustering algorithm to make the final decision in DSOM. The lowest rate of accuracy was around 96.5\% when they used 400 neurons. Where it was increased above 97\% when they used 900 and 1600 neurons. The false alarm rate also decrease with the increased number of neurons. In addition, the experimental results showed that the DSOM outperforms the single SOM in terms of the system overhead.

Braga et al. [162] introduced a lightweight DDoS flooding attack detection method based on SOM approach. They used 6 features namely: average of packets per flow, average of bytes per flow, average of duration per flow, percentage of pair-flows, growth of single-flows, growth of different ports. Their method showed high detection rate with low false positive rate. Jankowski and Amanowicz [163], on the other hand, compared different ML approaches for intrusion detection task in SDN paradigm. The experimental study showed that hierarchical learning vector quantization (Hierarchical-LVQ1) is more efficient than other approaches such as SOM and LVQ1.  

Nam et al. [164] used SOM for detecting DDoS attacks in SDNs. They used the following 5 features for the classification task: entropy of source IP address, entropy of source port, entropy of destination port, entropy of packet protocol, total number of packets. First, SOM was used for dimensionality reduction. Then, they proposed two classification techniques. The first one is based on an algorithm that combines SOM and k-NN. The second one is based on SOM with center-distributed classification, where in this case SOM is trained only on normal traffic samples. Then, the algorithm calculates the distance between each input sample to a universal reference point where the trained data is within a hyper-sphere, centred around that reference point with a predefined threshold. If the distance is less than that particular threshold, then it will be flagged as normal otherwise it will be flagged as anomalous. The traditional kNN achieved the highest accuracy but it showed high processing time. The first algorithm that combines SOM and kNN outperformed the second one (SOM distributed-center) in terms of detection rate and false positive rate. However, it showed a higher processing time in compared with the second one. SOM distributed-center showed a good detection rate with the lowest processing time however it has very high false positive rate.

\paragraph{Hidden Markov model (HMM) in SDN}\mbox{} \\
Fan et al. [54] investigated security situation assessment in SDNs based on an advanced HMM called multiple observations hidden Markov model (HMM). They used 12 observed features and each has two different values, where SVM is used to classify each feature value at different times into 1 or -1 values. According to the risk vector, the situation values change to show the security risk of SDN. They used Baum-Welch algorithm for training and Viterbi algorithm for predicting the network state. The experimental results showed that status of Switch compromised attack has the highest average value followed by ARP attack, OpenFlow flooding attack, scanning attack and normal situation respectively. The prediction accuracy was 88.25\%. Shan-Shan and Ya-Bin [165], on the other hand, introduced a model for detecting advanced persistent threat (APT) in SDNs based on HMM. APT is a multi-stage complex attack. Therefore, the authors proposed HMM to identify the stage of APT. Again, in this study, Baum-Welch algorithm was used for training and Viterbi algorithm for predicting the stage of APT. The experimental study showed that HMM was able to detect the stage of APT with low overhead.

\paragraph{Restricted Boltzmann machine (RBM) in SDN}\mbox{} \\
MohanaPriya and Shalinie [166] studied detection of DDoS attacks in SDN based on restricted Boltzmann machine.                                                                                                                                                                                                 RMB was trained using Constrastive Divergence (CD) algorithm. They used the following features as the input vector of RMB: source IP address, destination IP address, source port, destination port and protocol type. Their proposed model achieved a detection rate of 92\% with a false positive rate of 8\% based on the dataset generated by hping3 tool.

\paragraph{Unsupervised deep learning approaches in SDN}\mbox{} \\
Mao et al. [55] proposed an approach for construction of routing table based on deep belief networks. The experimental results showed that the proposed approach outperforms traditional open shortest path first (OSPF) protocol in terms of throughout and average delay per hope. In addition, the experimental results showed that the proposed routing method can run more than 100 times faster on a GPU than on a CPU. 

Zhang et al. [56] introduced a hybrid deep neural network for SDN-based network application classification, which consists of the stacked auto-encoder and softmax regression layer. SAE was used as unsupervised-learning-based feature extractor, whereas softmax regression was used as a supervised classifier. The proposed model achieved higher classification accuracy over SVM in terms of accuracy, precision, recall and F1-measure. 

Niyaz et al. [57] proposed a DDoS detection system based on SAE as feature extractor in SDN. A softmax layer is also used for classification. Their proposed system can identify DDoS attacks with an accuracy of 95.65\% for 8-class classification. It also achieved an accuracy of 99.82\% for binary classification to show whether an attack happened or not with very low false-positive (0.3\%). It also outperformed soft-max and neural network classifiers. Liu et al. [58] SAE was used for extracting spatio-temporal features of content popularity. Then they also used softmax classifier for the classification task. Their model achieved 2.1\%-15\% and 5.2\%-40\% accuracy improvements over neural networks and auto regressive, respectively.

\paragraph{Other unsupervised ML approaches in SDN}\mbox{} \\
Other unsupervised ML methods were used in the SDN paradigm for intrusion detection task. In this context, He et al. [167] introduced an anomaly detection and mitigation method based on a two-stage unsupervised learning approach for feature selection and clustering. For the first stage, they used Maximal Information Coefficient (MIC) to calculate the relation information between two continuous features and they calculate the relevancy, a symmetric uncertainty estimator, for discrete features. In the second stage, they applied a density peak clustering algorithm on the sampled data points. In the case of a hierarchy of SDN controllers is employed, then this approach can be used to reduce the volume of traffic shuffled across the network by locally analysing the traffic data in each controller. 

Ahmed et al. [168] proposed a method for mitigating domain name system (DNS) query-based DDoS attacks based on Dirichlet process mixture model for clustering traffic flows. The proposed system achieved better results over mean shift (MS) clustering approach.

\subsubsection{Reinforcement learning in SDN}\mbox{} \\
In this context, it is worth noting that Q-learning technique [74,129-133,last one] was widely applied in SDN paradigm for routing and adaptive video streaming. 

Al-Jawad et al. [169] proposed, LearnQoS, a RL-based framework that utilizes Q-learning for policy-based network management (PBNM) to optimize QoS requirements in multimedia-based SDNs. The RL was modelled by defining three elements: state, action, and reward. The state was represented by the traffic matrix. Four different actions were considered for the agent: 1) do nothing, 2) reduce data rate, 3) increase data rate, and 4) reroute. The rewards, on the other hand, were based on the service level agreements (SLA) requirements. In spite of the network overhead introduced by LearnQoS, the experimental results showed that the performance of the QoS was considerably improved when compared with the default multimedia-based SDN.

Sendra et al. [170] presented a routing method based on reinforcement learning approach. Given a set of possible paths and a set of network measurements (delay, loss rate and bandwidth), the agent tries to maximise the reward by choosing the path with less cost. Their method showed less loss rate and better jitter values, when compared to traditional OSPF routing protocol. Lin et al. 

[171] proposed QoS-aware adaptive routing (QAR) for distributed hierarchical control planes where Markov decision processes (MDP) with QoS-aware reward function used for modelling the system. Instead of using conventional Q-learning method, softmax action selection policy and state-action-reward-state-action (SARSA) approach were used for quality update.                                                                                                        

Kim et al. [172] investigated congestion prevention based on Q-learning approach. Compared with Dijkstra's algorithm and extended Dijkstra's algorithm, the proposed approach showed better results when the size of transmitted data increases.                                                                                                                                                                                                                                               

Uzakgider et al. [173] proposed an adaptive video streaming method based on Q-learning approach, in which Markov decision process (MDP) was used for modelling the system. The experimental results showed that their proposed system achieves better results when compared with the shortest path routing and greedy-based approaches.                                                                                                                                                                                                                     

Bentaleb et al. [174] proposed an end-to-end SDN-based intelligent architecture for large-scale HTTP adaptive streaming (HAS) systems, where partially observable Markov decision process (POMDP) was used for modelling the system. Then Q-learning based approach that employs a per-cluster decision algorithm was used to maximize the QoE. The experimental study showed that their system was able to increase the video stability and achieve better QoE fairness and network resource utilization. Jiang et al. [175] proposed, Q-FDBA, an on-line Q-learning-based dynamic bandwidth allocation algorithm for better QoE fairness. Q-FDBA showed better results when compared to bandwidth-aware (BA) streaming and QoE fairness framework (QFF). 

\begin{table*}[ht!]
\centering
\scriptsize
\caption{A summary for applying deep learning (DL) techniques to the SDN paradigm}
\begin{tabular}{|c|c|c|c|}
\hline 
Reference & \makecell{DL Approach} & Task & Findings \\ 
\hline 
Tang et al. [44] & GRU-RNN & Intrusion detection &  \makecell{Outperformed DNN [149], SVM and NB Tree\\with an accuracy of 89\%.} \\ 
\hline 
Mao et al. [55] & DBN & Routing & \makecell{Outperformed OSPF protocol in terms\\ of throughout and average delay per hope.} \\ 
\hline 
Zhang et al. [56] & SAE & \makecell{Feature extraction for \\network application classification} & \makecell{Outperformed SVM in terms of accuracy, \\ precision, recall and F1-measure.} \\ 
\hline 
Niyaz et al. [57] & SAE & \makecell{Feature extraction for\\ DDoS detection} & \makecell{An accuracy of 95.65\%, 99.82\% for 8-class and\\2-class classification respectively. Outperformed \\soft-max and neural network classifiers.} \\ 
\hline 
Liu et al. [58] & SAE & \makecell{Feature extraction for content\\popularity prediction in \\information centric network} 
& \makecell{Accuracy improvements over neural networks and \\auto regressive, 2.1\%-15\% and 5.2\%-40\% respectively.} \\ 
\hline 
Tang et al. [149] & DNN & Intrusion detection &  \makecell{Outperformed Naive Bayes, SVM and DT\\ with an accuracy of 75.75\%.}
 \\ 
\hline 
Lazaris and Prasanna [150] & LSTM-RNN & Time series traffic prediction & \makecell{ An average mean absolute percentage error (MAPE) \\ of 12\% for approximating real flow sizes \\ on CAIDA traces
and MAPE of 3.9\% on \\ simulating the network topology of Google's B4.} \\ 
\hline 
Azzouni and Pujolle [151] & LSTM-RNN & Traffic matrix prediction & \makecell{Outperformed an efficient dynamic routing \\ heuristic by finding the near \\ optimal path in shorter time using generated data.} \\ 
\hline 
Azzouni and Pujolle [152] & LSTM-RNN & Traffic matrix prediction & \makecell{Outperformed linear forecasting models \\ (ARMA, ARAR, HW) and feed \\ forward neural network (FFNN) using real data.} \\ 
\hline 
Huang et al. [153] & \makecell{Deep MLP (DNN),\\CNN and LSTM} & \makecell{Adversarial attacks on \\ SDN-based deep IDS} & \makecell{JSMA attack showed a significant impact on\\ the deep ML models ranges from 14 to 42\%,\\JSMA-RE reduced the accuracy of MLP to 35\%. \\
FGSM caused a significant reduction in \\ the accuracy of LSTM (more than 50\%).}\\
\hline 
Stampa et al. [179] & \makecell{Deep deterministic \\ policy gradients} & \makecell{Routing} & \makecell{Outperformed their primary benchmark.} \\ 
\hline 
Streiffer et al. [181] & \makecell{A3C} & \makecell{Automating data center \\ network topologies management} & \makecell{Finds near optimal solutions across a range of topologies.} \\ 
\hline 
\end{tabular} 
\end{table*}

Geng [176] presented MIND, which employs online version of Relative Entropy Policy Search using Reproducing Kernel Hilbert Space (REPS-RKHS) approach to learn the probability distribution of choosing the top-k best path.  REPS-RKHS is a combination of Relative Entropy Policy Search (REPS) and Reproducing Kernel Hilbert Space (RKHS) embeddings in order to provide a more stable, effective learning progress for non-parametric reinforcement learning. Online REPS-RKHS, on the other hand, is more appropriate for real world problems such as routing, which require a real time performance. MIND has a policy generation module choose an optimal routing policy by learning based on traffic data and network state.

Chavula et al. [177] used RL approach for SDN-based traffic engineering in UbuntuNet Alliance, which is the regional internetwork for National Research and Education Networks (NRENs) in southern and eastern Africa. The performance rewards were calculated based on distance (packet delay), available capacity on the link, and the resultant load (number of flows) at the next hop. The implementation of Q-learning approach consisted of two tables: (1) a local Q-values table at each network node and (2) a global aggregation table managed by the network controller. Each switch conducts QoS measurements on its next neighbors and passes them to the controller. The controller makes use of both active measurement data and interface-level statistics to update the Q-values based on calculating the reward values. The optimal path is selected by probabilistically choosing the forwarding link based on Q-values at each switch. They implemented the Q-learning approach using Mininet emulator to distribute traffic through multiple forwarding links in a way that maximizes the throughput and reduces the latency. For multipath configurations, the best values of latency and jitter were achieved when the rewards were based on both the available link capacity and latency. On the other hand, the best throughput was achieved when the rewards were based on the links' available bandwidth. The lowest values of latency were obtained with single path forwarding, where the rewards are based on the link delays. Also, single path forwarding achieves the lowest jitter.

Francois and Gelenbe [178] proposed Cognitive Routing Engine (CRE) for SDNs. CRE consists of 3 main modules. The first module is Cognitive Routing Algorithm Module (CRAM), which employs random neural network with reinforcement learning for finding network paths that maximize a customizable objective function to meet QoS requirements of host applications. Selection of the port that will be used as the next hop can be done in two different modes: (1) the exploratory mode and (2) the exploitation mode. In the exploratory mode, it chooses the output port randomly, but when CPN is in the exploitation mode, it will choose the neuron which has the highest probability. Accordingly, when the reward of the new path is higher than the threshold, then the path is considered valid and the corresponding weights are updated. Otherwise, RNNs have made the wrong decision, and therefore, the weights are updated to allow other paths to be selected. The second module is Network Monitoring Module (NMM), which obtains the network state information and notifies the CRAM. The last module is Path-to-OF Translator Module (PTM), which converts the paths found by CRAM into the appropriate OF messages. The experiment study, which has been conducted based on Mininet emulator and GÉANT network topology showed that CRE reaches near optimal paths with 9.5 times less monitoring data than conventional SDN. However, the solutions of CRE were on average 1.65\% worse than the optimal RTT.

Stampa et al. [179] investigated a deep RL approach for SDN-based routing optimization based on deep deterministic policy gradients approach [180]. The state of the deep RL agent is determined by the traffic matrix, the action by a tuple of link weights and the reward is based on the mean network delay. The proposed deep RL agent outperformed their primary benchmark. Streiffer et al. [181], on the other hand, introduced DeepConf, a novel RL-based SDN architecture for developing and training deep ML models for automating data center network topologies management. DeepConf consists of three components: (1) the network simulator for training Deep RL agents, (2) an abstraction layer to facilitate communication between the DeepRL agents and the network, and (3) the DeepRL agents, which encapsulate data center functionality. The authors used Asynchronous Advantage Actor Critic (A3C) approach which employs a deep network to approximate the policy and the value function. The agent's state space is the network topology whereas the action space for the model is represented by a vector which corresponds to different possible link combinations. The goal here is to maximize link utilization and minimize the average flow-completion time. The learning model makes use of a Convolutional Neural Network (CNN) to compute policy decisions, one CNN-block for each state space. The authors evaluated two clos-style data center topologies namely: Fat-tree and VL2 where the experiments showed that DeepConf was able to find near optimal solutions across a range of topologies. In this context, the research efforts made to apply deep learning techniques in SDN are briefly summarized in Table 3.


\subsubsection{Semi-supervised learning in SDN}\mbox{} \\
Semi-supervised learning [182-185] were also used in SDN, but much less common compared with other learning approaches. The research was focused on traffic classification [182, 184], routing [183] and intrusion detection [184]. Wang et al. [182] proposed a new framework for QoS-aware traffic classification based on semi-supervised learning. This approach can classify the network traffic according to the QoS requirements. The system allows achieving deep packet inspection (DPI) and semi-supervised learning based on Laplacian SVM. The Laplacian SVM approach outperformed a previous semi-supervised approach based on k-means classifier.                                                                                                                                                                                                                                                                                                               

Chen and Zheng [183] introduced an efficient routing pre-design solution based on semi-supervised approach. The study suggested using an appropriate clustering algorithm such as Gaussian mixture model and k-means clustering for feature extraction. Thereafter, a supervised classification approach such as extreme learning machine (ELM) can be used for flow demand forecasting. The authors also suggested using an adaptive multipath routing approach based on analytic hierarchy process (AHP) for handling to elephant flows according different constraint factor weights.                                                                                                                                                    

Li et al. [184] proposed a new method for fine-grained traffic classification based on semi-supervised approach called nearest application based cluster classifier (NACC). Unlike traditional methods, which use one feature vectors, this algorithm constructs a matrix with several cluster centroids based on k-means clustering to represent the application. The algorithm uses a small number of labelled flows to build a supervised dataset. Then collects the unlabelled flows to be merged with previously collected dataset, based on investigating the correlated flows, which is used to map an application to different clusters. The experimental results showed a good identification accuracy reaching 90\%. 

Wang et al. [185] introduced an intrusion detection method based on semi-supervised approach for wireless SDN-based e-health monitoring systems. The proposed system employed the concept of off-line training and on-line testing to allow running localized intrusion detection on wireless massive machine-type communications (mMTC) devices. Their system adopts semi-supervised learning on the basis of modified contrastive pessimistic likelihood estimation (CPLE), in which they replace the maximization calculation by a relaxation function. CPLE [186] performs semi-supervised parameter estimation for likelihood-based classifiers. The proposed modified CPLE outperformed Naive Bayes, SVM, DNN, self-training (semi-supervised approach) and the original CPLE based on the experiments conducted on NSL-KDD dataset.

\subsection{Meta-heuristic Algorithms Used in SDN}\mbox{} 
A large variety of meta-heuristic algorithms such as ant colony optimization [187-195], evolutionary algorithms [196-198], genetic algorithms [199-207], particle swarm optimization [208-213], simulated annealing [214-216], bee colony optimization-based [217,218], whale optimization [219,220], firefly optimization [221], bat algorithm [88], teaching-learning-based optimization [90] and grey wolf optimization [222] were used in SDN.

\subsubsection{Ant colony optimization in SDN}\mbox{} \\
ACO has been widely used for solving various networking problems such as routing [187-189,192], load balancing [190,191], network security [193,194] and maximizing network utilization [195]. Dobrijevic et al. [187] presented an ACO approach for flow routing. The proposed model also employs quality of experience (QoE) estimation models and attempts to maximize the user QoE for multimedia services. The experimental results showed promising QoE improvements in compared with shortest path routing.

Wang et al. [188] introduced two ACO-based algorithms for routing and spectrum assignment (RSA). The first one is ant colony optimization algorithm of minimum consecutiveness loss (ACO-MCL). The second one is ant colony optimization algorithm of maximum spectrum consecutiveness (ACO-MSC). The experimental results show that the proposed algorithms can reduce the blocking rate by at least 5\% and perform better in spectrum efficiency.

Gao et al. [189] presented, CACO-RSP, a traffic engineering ACO-based approach to solve the routing rule space occupation problem for multiple unicast sessions. CACO-RSP algorithm considers both the local and global pheromone trail to guide the searching. The simulation results show that the proposed algorithm was successful in reducing the routing rule consumption.

Di Stefano et al. [190] proposed, A4SND, a load-balancing algorithm with low complexity and hight scalability based on extending AAA algorithm, which is a modified version of ACO algorithm. A4SDN is different from the ACO-based routing algorithms in the following two points. First, the opposite interpretation of the pheromone trails. Second, the sub-path pheromone evaluation. The decision making policy in A4SND is based on pheromone laid on sub-paths, which makes it appropriate for on-line decision problems such as the routing. A4SND was emulated with the Mininet tool with two well-known variants of the Dijkstra's shortest path algorithm. The experimental results showed that A4SDN was able to outperform other two variants of Dijkstra's shortest path algorithm in terms of network throughput, end-to-end latency and packet loss.

Wang et al. [191] presented, LLBACO, a link load balancing algorithm based on ACO. LLBACO considers link load as main factor, whereas delay and pack-loss are the secondary factors. The experimental results showed that LLBACO has a better performance in balancing the network load and improving the QoS. 

Parsaei et al. [192] investigated a method for providing QoS for remote telesurgery applications. The proposed approach periodically collects statistics of network state and uses ACO for computing the best path between surgeon and patient. The proposed approach improved the average End-to-End delay, packet loss ratio and peak signal to noise ratio (PSNR) with values of 56.3\%, 50.5\% and 1.33\% respectively. Chen et al. [193] proposed an ACO-based method for detecting low-rate distributed denial of service (LDDoS). The proposed method consists of three stages, an information heuristic stage, a multi-agent ACO-based algorithm, and  backward and forward search stages. The experimental results showed that the detection rate was more than 89\% and the accuracy is greater than 83\%.

Liu et al. [194] introduced a defence mechanism of random routing mutation (RRM) in SDNs. The authors constructed an entropy matrix of network traffic characteristics and detected network anomalies. Routing mutation is triggered according to the anomaly detection results. The generation of a random routing path is formalized as a 0-1 knapsack problem, which is calculated by means of an enhanced version of ACO algorithm. The experimental results showed that the proposed approach was successful in defending against reconnaissance, eavesdropping, and DoS attacks. The main drawback of this approach was the increased burden on the controller. Yi et al. [195] used  WiseAnt Colony Optimization (WACO) algorithm to solve the Bounded Forwarding-Rules Maximum Flow (BFR-MF) problem (i.e. limited resource problem in TCAM-based SDN switches). The experimental results showed that WACO can improve the network utilization and throughput by up to 16\% on the premise of a certain level of QoS.

\subsubsection{Evolutionary algorithms in SDN}\mbox{} \\
The evolutionary algorithms were employed for solving controller placement problem in distributed SDNs [196], routing [197] and moving target defence [198]. 

Zhang et al. [196] investigated the problem of controller placement by taking into consideration controller-to-switch and controller-to-controller delays for wide area networks, where they proposed two evolutionary algorithms. The first one, denoted as Evo-Place, finds a set of Pareto controller placements. The second one, denoted as Best-Reactivity, finds the final placement that minimizes the average reaction time perceived at the switches. The experimental study showed that a better Pareto frontier obtained by Evo-Place compared to Rnd-Place (a basic randomized algorithm), given the same number of considered placements and the same number of iterations. 

Fernandez-Fernandez et al. [197] presented an SDN-based routing strategy that considers both QoS requirements and energy awareness. They employed a Multi-Objective Evolutionary Algorithms (MOEA) based on the Strength Pareto Evolutionary Algorithm 2 (SPEA2). They employed two objective functions. The first one is based on the performance requirements for the communications between the control and data planes. The second objective, related with energy awareness, aims at minimizing the number of links that need to be activated when a connection request arrives. The proposed routing algorithm significantly outperformed a modified shortest path routing in terms of accepted demands.

Makanju et al. [198] suggested using evolutionary computation (EC) techniques for moving target defence in SDN. Moving target defence (MTD) systems have three main challenges: 1) how to choose another configuration, 2) adapting to next configuration, and 3) when to start the adoption. They mentioned that a multi-objective genetic algorithms (MOGA) can be suitable for MTDs requirements.

\paragraph{Genetic algorithms in SDN}\mbox{} \\
Genetic algorithm were mainly used for routing [199], virtual network planning [200], optimizing the cost of deploying deep packet inspection functions in SDN [201], load balancing [202-204], designing optimal observation matrices [205], resource reallocation for SDN-based data centres [206], and congestion avoidance [207].

Yu and Ke [199] introduced, GA-SDN, a genetic algorithm-based routing algorithm for enhancing the video delivery quality over SDNs. The experimental results showed that GA-SDN outperformed Bellman-Ford routing algorithm in terms of packet drop rate, throughput, and average peak signal-to-noise ratio. Wang et al. [200] studied the problem of virtual network planning in SDNs based on GA approach. The virtual network refers to customise a network topology and place the controllers that should meet a given QoS requirements. Their proposed approach outperformed a greedy solution. In addition, compared to k-medoids, the proposed algorithm showed that it can reduce the number of required SDN controllers. However, k-medoids showed that it can outperform the GA-based solution in terms of the average latency.

Bouet et al. [201] presented a cost-based method for optimal deployment of deep packet inspection (DPI) engines in NFV-SDNs. The experimental results showed that the method was able to reach a trade-off between the number of DPI engines and network load. In addition, the global cost can be reduced up to 58\% when relaxing the constraint on the used link capacity. Chou et al. [201] introduced a policy-based load balancing system, where the authors proposed a genetic-based load balancing approach. Their system outperformed load-based, round-robin, and random choice-based approaches in terms of arithmetic average for coefficient of variation. 

Kang and Kwon [203] also introduced an SDN-based load balancer where the authors introduced a mapping solution as a tree structure. The authors suggested that a master (super) controller, which should be responsible for the load balancing operation can be represented as a root node, and the other controllers can be indicated by an internal node and finally a switch will be represented by a leaf node. Mahlab et al. [204] proposed a fragmentation-aware load-balancing strategy for optical networks. The goal here is to optimize the fiber-load across the network.  The authors employed an entropy-based metric for measuring the load imbalance and then used it for designing a joint entropy/hits utility function for the optimization, which was solved by a GA. 

Malboubi et al. [205,206] presented, SNIPER, a framework for designing the optimal observation matrix that can lead to the best estimation accuracy using matrix completion techniques. The authors used GA and PSO to deal with large-scale optimal observation matrices. The experimental results showed that the proposed approach can be applied to many network monitoring applications in large-scale networks under hard resource constraints. For instance, the authors showed that by measuring only 8.8\% of all per-flow path delays in Harvard network [1], congested paths can be detected with probability of 0.94. 

Tajiki et al. [207] introduced, CECT, a congestion avoidance scheme for SDN-based cloud data centres. They used a routing architecture to reconfigure the network resources, which is triggered when a predefined time interval or when a congestion occurs. They proposed a meta-heuristic approach based on genetic algorithms to find the optimal solution for the optimization problem. CECT can enhance the total network throughput up to 3x while it decreases the packet loss up to 2x when compared with ECMP (Equal Cost Multiple Path).

\subsubsection{Particle swarm optimization in SDN}\mbox{} \\
Used for routing [208, 209], resource management [210], multi-tenant virtual network customization [211], multi-class routing [212], and detection of DDoS attacks [213].

Awad et al. [208] proposed, PSOPR, a PSO-based power-efficient routing approach for solving the problem of flow table overflow. The experimental results showed that PSOPR can achieve more than 90\% of the optimal network power consumption while it requires only 0.0045\% to 0.9\% of the optimal computation time in real-network topologies obtained from SNDlib (a library of test instances for Survivable fixed telecommunication Network Design). Moreover, PSOPR resulted in shorter routes than the optimal routes obtained by the optimization package CPLEX.

Xiong et al. [209] proposed, PALM, power-aware light-path management algorithm for traffic prediction in SDN-based elastic optical networks. The goal here is to avoid tearing down a light-path that becomes idle and takes into consideration reducing the switching power, which is required to re-establishment a light-path again in short time later. They also introduced a PSO-BP neutral network model to assist the PALM algorithm in accurately predicting future traffic demands and reducing the power consumption. More precisely, a PSO-BP neural network is used to predict the traffic load for every light-path within a particular time interval. PLAM achieved the power saving of 36\% and 16\% when compared with energy-efficient many-cast (EEM) algorithm and dynamic scheduling and distance-adaptive transmission (DS+DAT) algorithm respectively.

Chang et al. [210] used GA and PSO for network allocation in 5G under NFV/SDN architecture. In terms of energy consumption GA-PSO approach achieved better results over both greedy strategy and OSFP approaches. In heavy load network environments, the proposed system can save energy nearly about 32\% less than the OSPF. Their experimental results showed also a larger average link resource utilization in compared with OSPF and greedy strategy. Li et al. [211] introduced a PSO-Based virtual SDN customization for multi-tenant cloud services. Their experimental results showed that the PSO algorithm can significantly improve utilization rate of the underlying network bandwidth.

Abdulqadder et al. [212] presented SecSDN-cloud, an integrated, secure, SDN and cloud-based architecture for providing enhanced quality of service (QoS) and more secure network. In SecSDN, the authors introduced an enhanced PSO multi-class (E-PSO) routing protocol, which takes three parameters into consideration namely: node congestion, link congestion and delay. The proposed PSO-based multi-class routing protocol included a data traffic classifier per-node to segregate traffic flows into QoS classes. Then it moves to the route selection process in which a node to check the QoS of other nodes. Another node is selected as the one hop neighbour for transmission, if it has a higher QoS value than the current node. SecSDN showed better results in terms of throughput when compared with OpenSec and AuthFlow. SecSDN showed better results in terms of throughput when compared with OpenSec and AuthFlow. In addition, it achieved better end-to-end delays when compared with OMC-RPL and a lower packet loss rate when compared with RPL based SDN and FlowDefender. 

Dayal and Srivastava [213] proposed a model for DDoS detection based on RBF-based neural network with PSO optimization. The author used the following features: average packets per flow, average bytes per flow, number of flows per second, average duration per flow, entropy of destination IP addresses per second, entropy of source IP address per second and entropy of IP protocol per second. The experimental results showed RBF-PSO achieved better accuracy and less training time when compared with NN-BP and NN-PSO.

\subsubsection{Simulated annealing in SDN}\mbox{} \\
Simulated annealing (SA) used for solving controller placement problem [214,215] and dynamic controller provisioning [216].                                                                                                                                                                           Hu et al. [214] proposed a method for solving the problem of placing controllers in SDNs in order to maximize the reliability of control networks. They proposed a novel metric that indicates the reliability of the SDN control network, called expected percentage of control. The aim of the optimization is to minimize the expected percentage of control path loss. The simulation results showed that SA achieved the best results when compared to other methods. However, the results showed that adding a large number of controllers has an adverse effect. The reason for that is adding more additional controllers will reassemble a full mesh network with too many control paths leading to a low reliability [214,215]. 

Bari et al. [216] used SA for deploying multiple controllers within an WAN. In terms of flow set-up time, the experimental results showed that SA can outperform both greedy knapsack (GK) and, a single controller for the entire network called 1-CRTL. However, SA required much longer time to run than GK. On the other hand, the overhead of SA is smaller than GK due to the fact that it employed fewer number of controllers and which is very close to 1-CTRL.

\subsubsection{Bee colony optimization-based algorithms in SDN}\mbox{} \\
Mohammadi and Javidan [217] used ABC algorithm for traffic engineering in SDN-based video surveillance systems.                                                                                                                                                                                                    The main goal of this work was to increase the quality of the received video data at monitoring server, which were streamed by cameras. The performance of the proposed method has been compared to Dijkstra routing algorithm in different scenarios. The experimental results showed that the proposed method has achieved better performance compared to Dijkstra algorithm in terms of average end to end delay, packet delivery ratio and PSNR. 

Kang and Choo [218] proposed a system for load balancing in SDN-based cloud systems based on centralized dynamic load balancing approach, which uses the history from each cloud in to reach its decision. The algorithm includes two steps: job selection and job dispatching. It adopts Best-Fit approach, which means that it will dispatch a job to the shortest average response time (ART) among available clouds. The authors compared the performance of their proposed approach to honeybee foraging algorithm (HFA) and round robin (RR) approaches. The experiments showed that the average response time of the proposed system has a higher saturation point than the other two approaches.

\subsubsection{Whale optimization algorithm in SDN}\mbox{} \\
Farshin and Sharifian [219] proposed, MAP-SDN, a framework for assignment and provisioning for SDN-based cloud data centers with different classes of service. The authors compared the performance of WOA with PSOGSA algorithm. PSOGSA, combines the exploitation of particle swarm optimization (PSO) with exploration of gravitational search algorithm (GSA). The experimental results showed that WOA was more accurate and it can reach better results with lesser computation time.

WOA also has been used for achieving clustering-based routing for heterogeneous, randomly distributed and dense IoT networks [220]. The proposed approach consisted of two stages: set-up stage and transmission stage. At the set-up stage, the SDN controller divides the sensing area into virtual zones for balancing the number of cluster heads (CH). Then at the transmission stage the controller uses WOA for each virtual zone to determine the optimal set of CHs in that particular virtual zone. The simulation results showed that the proposed approach can minimize the power consumption in compared to conventional routing protocols. In addition, the total number of packets received by the sink throughout the simulation time has improved by approximately 55\% in compared to conventional clustering protocols, and 20\% compared to the optimization based (PSO) clustering protocol.

\subsubsection{Firefly optimization in SDN}\mbox{} \\
Sahoo et al. [221] used PSO and FireFly algorithm (FFA) for solving the controller placement problem in SDN-based Wide Area Network (WAN). They considered three metrics namely: 1) controller to switch latency, 2) inter-controller latency and 3) multi-path connectivity between the switch and controller. In addition, they presented Average Delay Rise (ADR) metric to measure the increased delay due to the failure of the primary path. The experimental results showed that FFA produced efficient and accurate results.

\subsubsection{Bat algorithm in SDN}\mbox{} \\
Sathya and Thangarajan [88] employed binary bat algorithm (BBA) for feature selection in an SDN-based intrusion detection system. Based on the experiments conducted on NSL-KDD dataset, the selected features achieved a detection rate of 90.9\%, 91.1\%, 80.2\% and 98.1\% for DoS, Probe, R2L and U2R attack types respectively with low false positive rates when used with J48 decision tree.

\subsubsection{Teaching-learning-based optimization in SDN}\mbox{} \\
Mohammadi and Javidan [90] used TLBO for providing SDN-based QoS for remote telesurgery applications. The proposed model calculates the optimal path between the surgeon and operating room based on network resources status. TLBO achieved the best PSNR, end-to-end delay, packet loss ratio and SSIM when compared to the shortest based approach.

\subsubsection{Grey wolf optimization in SDN}\mbox{} \\
Farshin and Sharifian [222] proposed PSO-based chaotic Grey Wolf Optimizer algorithm for dynamic controller allocation in a SDN cloud-based 5G cellular networks. They used chaotic maps to avoid local optimum and improve the convergence speed. Due to the fact that the proposed framework consisted of several classes and each class included its own switching topology and controllers, we should use two nested chaotic GWO. The experimental results showed that chaotic GWO can achieve results that are more accurate (1\% better) during fewer iterations (68\% less) compared to the traditional PSO.

\subsection{Fuzzy Inference Systems in SDN}
Fuzzy inference systems were also widely used in SDN paradigm [223,225-229]. The main research was focused on introducing new protocols [223], intrusion detection [225-227], selection the optimal network deployment schemes [228] and traffic engineering [229]. 

Abdolmaleki et al. [223] proposed a topology discovery protocol for software defined wireless sensor networks. Their proposed protocol showed better performance when compared with software defined networking solution for wireless sensor networks (SDN-WISE) [224]. Dang-Van and Truong-Thu [225] presented an approach for preventing DDoS attacks based on Sugeno's fuzzy inference method. They analysed a real network traffic to conclude DDoS indicators and thresholds. The input of the fuzzy module were the following two parameters: rate of packets having inter arrival times in range (0-0.2 ms] and rate of flows having only one packet per flow. The experimental results showed the ability to detect and filter 97\% of attack flows. In addition, it achieved a reasonably low false positive rate (approximately 5\%) and maintained a reduction of flow entries of 50\% during attack time.

Rezaei et al. [226] proposed a cooperative defence approach against DoS flood attacks based on fuzzy modelling. The model has two parameters, which are trust in service provider and service sensitivity. The proposed system achieved lower computational load and response time compared to SynCookie and service provider's firewall. Dotcenko et al. [227] proposed an information security management system based on Mamdani's fuzzy inference approach. Threat degree was determined by the fuzzy inference module, which mainly depends on threshold random walk with credit-based rate limiting (TRW-CB) and rate limiting algorithms. The proposed system was able to detect 95\% of the attacks with 1.2\% false positive rate. 

Abdallah et al. [228] studied the selection of network deployment schemes based on Mamdani's approach. Their model consists of 21 rules and has the following 5 input parameters:  network size (NS), the network utilization (NUT), network updates (NUP), initial technology deployment (ITD), ease of management (EM). The output parameter is the network technology choice (NT). This model can help the operators, who provide different options of networks to deploy. Mohammadi and Javidan [229] proposed a traffic engineering method for SDN-based video surveillance systems using adaptive,type-2 fuzzy approach, which showed better results when compared with type-1 fuzzy approach, available bandwidth based traffic engineering approach and, OSPF routing protocol in terms of end to end delay, peak signal to noise ratio (PSNR) and packet loss ratio.

\subsection{Other AI-based Approaches in SDN}
Other researchers proposed many hybrid models that make use of two or more intelligent approach to enhance the overall performance and accuracy of these algorithms. 

Petrangeli et al. [230] investigated preventing video freezes in HTTP adaptive streaming based on combination of random under-sampling boosting (RUSBoost) algorithm and fuzzy logic approach. The Experimental results showed that the proposed system was able to reduce video freezes with 65\% and freeze time with 45\%, when compared with fair in-network enhanced adaptive streaming (FINEAS) and Microsoft ISS smooth streaming (MSS).                                                                                                                                                                                      

De Assis et al. [231] proposed a method for mitigating DoS/DDoS attacks based on a game theatrical approach that employs Holt-Winters and genetic algorithm with fuzzy logic. Holt-Winters for digital signature (HWDS) approach was used for detecting and identifying anomalies and game theory decision-making (GTDM) approach was used for choosing the optimal defense strategy. HWDS was compared with fuzzy genetic algorithm digital signature (Fuzzy-GADS) approach where HWDS fared better in performance tests. Fuzzy-GADS, on the other hand, found to be more efficient on detecting the occurrence of stealthier anomalies.

Li et al. [232] introduced an intelligent hybrid intrusion detection system for SDN-based 5G network where random forest approach was used for feature selection and k-means++ with Adaboost approach was used for flow classification. The algorithm was able to achieved a good precision (94.48\%), recall (92.62\%) and f1-score (91.02\%) with a low false positive rate (0.54\%) based on 23 selected features from KDD99 dataset. Da Silva et al. [233] proposed an anomaly detection and classification method based on a two-phases approach: 1) lightweight phase based on entropy analysis and 2) heavyweight approach based on a hybrid ML approach, which employs k-means and SVM. Similar flows were clustered together where each cluster represents a particular traffic profile then the SVM algorithm is used to classify the flows in each cluster. The SVM showed an accuracy of 88.7\% and a precision of 82.3\%.

Sabih et al. [234] studied optimizing the performance of SDN based on hybrid intelligence approach, which makes use of a neural network model. GA and PSO algorithms were used separately to select the optimal set of inputs that maximizes the network efficiency. The experimental results showed that PSO achieves a better performance and a faster convergence in compared with GA. Li et al. [235] presented an intelligent SDN-based video management system, which uses face detection and recognition techniques based on EigenFace algorithm principal component analysis (PCA), FisherFace algorithm based on linear discriminant analysis (LDA), and local binary patterns (LBP).

Yuan e al. [236] investigated revenue maximization in data centers based on workload-aware SDN controller, which determines the optimal combination of a VM and routing path for each application. The proposed system uses a hybrid approach named hybrid chaotic simulated-annealing PSO (HCSP) that combines chaotic PSO and SA algorithm. Compared with OSPF and Round Robin approaches, the proposed approach was successful in reducing the RTT and increasing the revenue. Huang et al. [237] proposed a method for application identification based on DNS responses inspection and ML. A classification system, which adopts a voting strategy, was implemented by Weak Java API, from which three algorithms were used namely: random forest, rotation forest and random committee with random tree. The experimental results showed that the combination of DNS responses inspection and ML techniques was able to achieve higher accuracy compared to standalone ML techniques. Pillutla and Arjunan [238] introduced, FSOMDM, a fuzzy self organizing maps-based DDoS mitigation approach for cloud-based SDNs. FSOMDM, enhances the neural network approach by replacing the neurons of the traditional Kohonen neural network model by updating fuzzy rules. The experimental results showed that FSOMDM achieved a true positive rate (TPR) of 94\%.

Aibin et al. [239] employed Monte Carlo Tree Search (MCTS) for advanced traffic predication for bandwidth-on-demand services and provisioning of dynamic lightpaths based on Cross-Stratum Optimization architecture. MCTS builds a sparse search tree and chooses actions using Monte Carlo sampling approach. They compared the performance of the different approaches for selecting a data center that can be used for providing services to a request. These approaches are: the nearest, the cheapest, the least utilized data center (DC), hybrid assignment (combining the three simple approaches) and traffic predication approaches based on MCTS. In the experimental studies they considered data center models and pricing structure provided by Amazon Web Services. The hybrid assignment and traffic prediction were the two best approaches among the others. The traffic prediction approach resulted in low request blocking percentage and more efficient utilization network resources. Hybrid approach was slightly lower than the cost of using the MCTS approach. However, it showed a higher request blocking percentage in case of high traffic loads.

Gao et al. [240] presented a hybrid approach that combines supervised and unsupervised approaches for defending against Packet-In based flooding attacks. Their proposed approach consisted of two stages. In the first stage, Bayesian Network, which is a supervised learning algorithm, is used for classifying the potential compromised switches. In the next stage, they make use of fuzzy c-means approach to determine whether the packet-In messages flooding attack happened or not. They conducted their experimental study based on Mininet where they used Distributed Internet Traffic Generator (DITG) to generate the legitimate traffic and DARPA intrusion detection data sets were used as the malicious traffic. Based on the receiver operating characteristic (ROC) curves, we can observe that the system can achieve a good accuracy, however it is combined with relatively high false alarm rate. For instance, when the detection rate is 95.68\%, the false positive rate reaches 10\%. The system however achieves a low overhead, as it only needs to monitor the vulnerable switches rather than all the available switches.

\section{Conclusion}
In this paper, we provided a state-of-the-art overview of research efforts made for applying artificial intelligence techniques in SDN paradigm. Compared to our recent paper [13], this study showed an increased adoption of various AI techniques to solve a wide range of networking problems and address new challenges introduced by the SDN paradigm. This study showed that ML, meta-heuristics and fuzzy systems were the most common used AI approaches in SDN. In addition, it has shown that the recent advances in ML techniques such as deep learning and hybrid AI approaches can provide better results in compared with traditional ML approaches. Overall, AI approaches have been proved to be very useful tools in SDNs. However, more efforts towards studying the robustness of AI approaches under adversarial settings need to be take into consideration, as well.

\section*{Acknowledgment}
We would like to thank the anonymous reviewers for their insightful comments and constructive suggestions, which significantly  helped us to improve the quality of this work.

\end{document}